\documentclass[acmsmall,screen=true,anonymous=false,bookmarks=false]{acmart}
\usepackage{lipsum}
\usepackage{graphicx}
\usepackage{amsmath}
\usepackage{footnote}
\usepackage{mathtools}
\usepackage{mathrsfs}     
\usepackage{comment}
\usepackage[subrefformat=parens,labelformat=parens]{subfig}
\usepackage{bm}
\usepackage{multirow}
\usepackage{threeparttable,booktabs}
\usepackage{blkarray}
\usepackage{tikz}
\usetikzlibrary{positioning,calc,fit,decorations.pathmorphing,shapes.geometric,shapes.gates.logic.US,calc}
\usepackage{balance}
\usepackage{courier}                                       
\usepackage{cleveref}                                      
\Crefformat{figure}{Fig.~#2#1#3}                           
\Crefname{subfigure}{Fig.}{Figs.}
\Crefname{figure}{Fig.}{Figs.}
\usepackage[mathcal]{eucal}
\usepackage[]{algpseudocode}                               
\algrenewcommand\textproc{\texttt}
\makeatletter\let\float@addtolists\relax\makeatother
\usepackage{algorithm}
\usepackage{filecontents}                                  
\usepackage{pgfplots}
\usepackage{pgfplotstable}
\pgfplotsset{compat=newest}
\usepackage[figuresright]{rotating}
\usepackage{xcolor,colortbl}

\renewcommand{\vec}[1]{\boldsymbol{#1}}

\newcommand{\revise}[1]{#1}

\theoremstyle{plain}

\theoremstyle{definition}

\newtheorem{myproblem}{\textbf{Problem}}

\algrenewcommand\textproc{\texttt}

\usepackage[skip=1pt]{caption}            
\definecolor{CUHKorange}{RGB}{244,106,18} 
\definecolor{CUHKblue}{RGB}{0,111,190}    
\definecolor{CUHKgreen}{RGB}{0,127,128}   
\definecolor{CUHKred}{RGB}{228,46,36}     
\definecolor{CUHKyellow}{RGB}{198,148,34} 
\definecolor{CUHKdark}{RGB}{114,44,114}   
\definecolor{CUHKmiddle}{RGB}{144,44,144} 
\definecolor{CUHKorange}{RGB}{244,106,18} 
\definecolor{CUHKblue}{RGB}{0,111,190}    
\definecolor{CUHKgreen}{RGB}{0,127,128}   
\definecolor{CUHKred}{RGB}{228,46,36}     
\definecolor{CUHKyellow}{RGB}{198,148,34} 
\definecolor{CUHKdark}{RGB}{114,44,114}   
\definecolor{CUHKmiddle}{RGB}{144,44,144} 
\definecolor{CUHKlight}{RGB}{167,44,167} 
\definecolor{CUHKpurple}{RGB}{117,15,109}
\definecolor{CUHKgold}{RGB}{221,163,0}
\definecolor{CUHKribbon}{RGB}{244,223,176}
\definecolor{CUHKblack}{RGB}{34,24,21}
\usepackage{tcolorbox}
\tcbuselibrary{skins,breakable}
    {\endtcolorbox}

    {\endtcolorbox}

\graphicspath{{./fig/}}
\setcopyright{none}
\acmJournal{TODAES}
\usepackage{pgfplots}
\usepackage{mathtools}
\usepackage{url}

\usepackage[export]{adjustbox}
\usepackage[inline]{enumitem}

\definecolor{mygreen}{RGB}{211,237,194}    
\definecolor{myblue}{RGB}{177,199,221}     
\definecolor{myorange}{RGB}{235,200,141}   

\newcommand{\minisection}[1]{\vspace{.1in}\noindent{\textbf{#1}}.}

\begin{document}

\date{}

\title{
    MacroAgent: Regularity-Aware Macro Legalization with LLM-Agent-Designed Contour Algorithms
}

\author{Jiaxi Jiang}
\affiliation{
    \institution{The Chinese University of Hong Kong}
}


\author{Xufeng Yao}
\affiliation{
    \institution{The Chinese University of Hong Kong}
}

\author{Yuxuan Zhao}
\affiliation{
    \institution{The Chinese University of Hong Kong}
}

\author{Yuntao Lu}
\affiliation{
    \institution{The Chinese University of Hong Kong}
}

\author{Peiyu Liao}
\affiliation{
    \institution{The Chinese University of Hong Kong}
}

\author{Zuodong Zhang}
\affiliation{
    \institution{Peking University}
}

\author{Yibo Lin}
\affiliation{
    \institution{Peking University}
}

\author{Bei Yu}
\affiliation{
    \institution{The Chinese University of Hong Kong}
}

\renewcommand{\shortauthors}{Jiaxi Jiang et al.}

\begin{abstract}
Macros constitute a large part of the core area in modern very large-scale integration (VLSI) designs. Moreover, macro positions have a significant impact on the final quality of result (QoR), and macro legalization is typically the final step in determining the macro positions. However, existing approaches related to macro legalization either lack robustness or incur substantial computational costs or neglect the regularity between macros. To address these limitations,  we introduce MacroAgent. The novel framework is a four-stage approach: clustering, contour generation, template matching, and inter-cluster refinement. We propose leveraging Large Language Models (LLMs) to discover multiple, effective heuristic regularity-aware contour algorithms.  This framework successfully generates robust and effective algorithmic solutions for macro legalization. Compared with state-of-the-art macro legalization works, experimental results on TILOS and Chipyard benchmarks demonstrate a 2 to 8 fold improvement in layout regularity, a 3\% to 5\% reduction in routed wirelength with comparable congestion after global routing, and significantly better robustness with an acceptable runtime. \revise{Furthermore, end-to-end evaluation through Cadence Innovus place-and-route confirms that the regularity improvements translate into tangible PPA gains, including 2.9\% lower routed wirelength and 68.3\% TNS improvement over the DREAMPlace macro legalization baseline; it also achieves 1.8\% lower routed wirelength when integrated into the Innovus macro placement flow.}
\end{abstract}




\maketitle
\thispagestyle{plain}
\pagestyle{plain}

\section{Introduction}
\label{sec:intro}

Macros in very large-scale integration (VLSI) circuits, such as memory and spatial arrays, are pre-designed giant functional blocks. They have fixed dimensions (e.g., height and width) and optimized structures. 
Macro placement heavily constrains subsequent standard-cell optimization, which in turn indirectly impacts power, performance, and area by affecting the routed wirelength~\cite{mclellan2020innovus}. Macro legalization represents the final phase that determines legal macro positions at the end of macro placement process.

Traditionally, the physical design procedure for VLSI circuits is composed of multiple stages, including floorplan, macro placement, standard-cell placement, and routing. Macro placement occupies an early position in the design process and has a significant influence on subsequent phases~\cite{gao2022congestion}. Moreover, due to the large size of macros, minor position adjustments can significantly impact downstream optimizations~\cite{pu2024incremacro}. 

The research literature categorizes macro placement into two main categories. The first is mixed-size global placement (placing macros and standard-cells together, rough locations), then macro legalization (overlap removal)~\cite{lin2018regularity,agnesina2023autodmp,peng2023pplace,chen2023stronger,xuereinforcement,jiang2025regplace,pu2024incremacro}. During the first step, overlap constraints are relaxed, multiple objectives are considered. Then macro legalization removes remaining overlaps. The second category utilizes layout data structures to \revise{directly} place macros, optimize customized objectives via simulated annealing~\cite{chen2007mp, chang2017novel, vidal2019rtl, kahng2022rtl, kahng2023hier}. The first category is faster than simulated annealing in the second, which is less scalable~\cite{innovus_manual,agnesina2023autodmp}. 
In industry practice, macro placement relies on both senior engineers' expertise (e.g., manual placement) and algorithmic efforts~\cite{agnesina2023autodmp}. Complex projects may take hours to days~\cite{agnesina2023autodmp}. The industrial tool, Innovus, offers a flow similar to the first macro placement method in the research literature, with two commands: \texttt{place\_design -concurrent\_macro} (mixed-size global placement) and then \texttt{refine\_macro\_place} (macro legalization)~\cite{innovus_manual}. 

\begin{figure}[tb!]
    \centering
    \includegraphics[width=\linewidth]{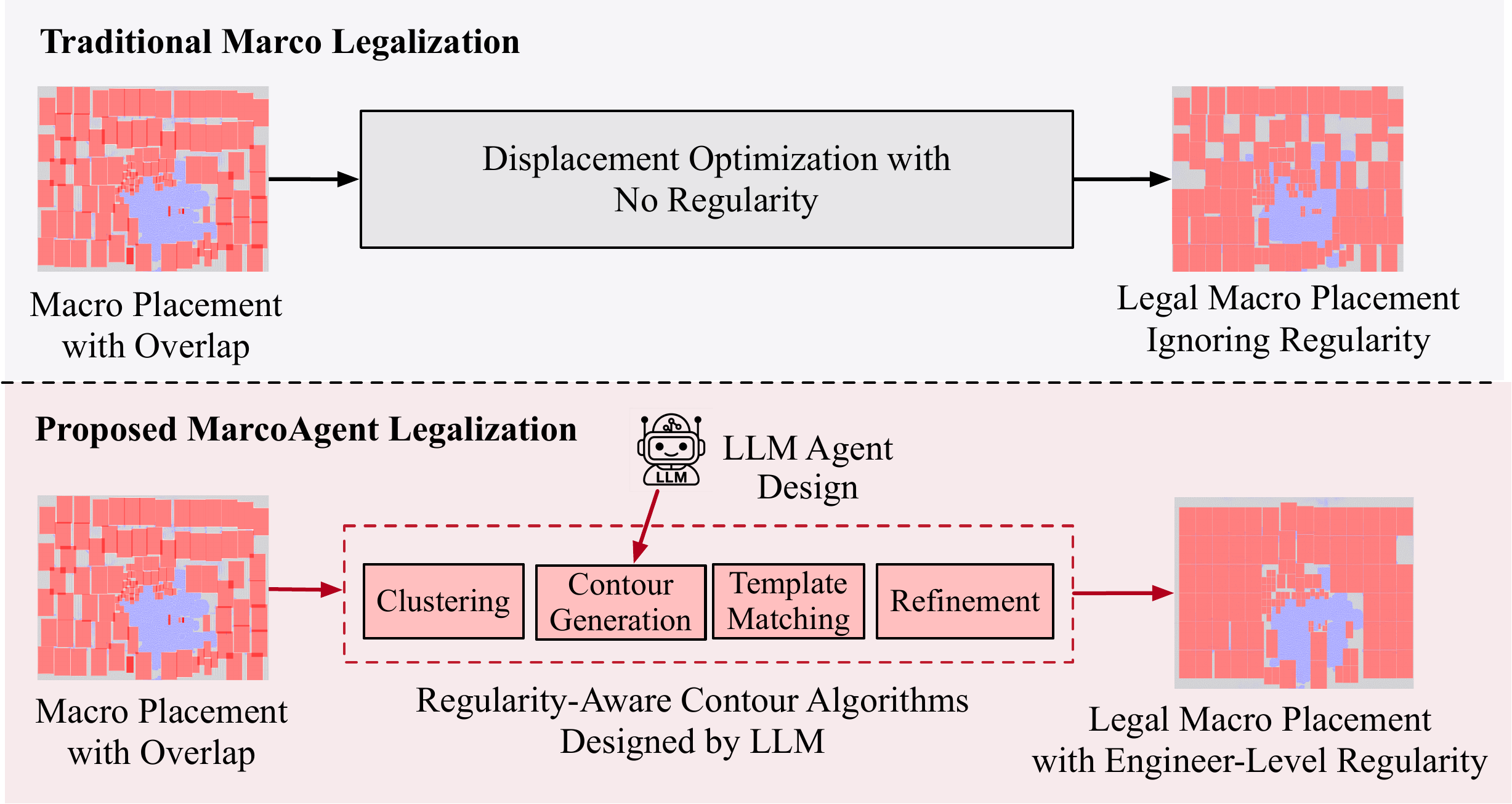}
    \caption{The MacroAgent framework consisting of four-stage legalization steps. MacroAgent achieves engineer-level regularity macro legalization results by regularity-aware contour algorithms designed by the LLM agent.}
    \label{figs:introduction}
\end{figure}

\minisection{Current Approaches to Macro Legalization}
Most prior works in macro legalization only attempt to minimize the displacement (i.e., the movements of macros). \cite{cong2008robust} proposes a legalization scheme that computes macro coordinates through the iterative adjustment of two constraint graphs, minimizing displacement by solving the linear programming problem.
\cite{lin2018regularity} extracts search points and feasible regions using the Puzzle algorithm, and iteratively places macros.
DREAMPlace 2.0 \cite{lin2020dreamplace} combines the two techniques mentioned above and has been applied in~\cite{agnesina2023autodmp,jiang2025regplace,pu2024incremacro}. The work in \cite{peng2023pplace} designed an occupancy-aware macro legalization algorithm that iteratively eliminates overlaps via heuristic approaches. The study in \cite{chen2023stronger} employs integer linear programming or simulated annealing to adjust macro positional relationships when initial sequence pairs prove infeasible.
Although these techniques can improve the robustness of macro legalization, they still fall short in two aspects: \textbf{first, they overlook the regularity between macros; second, they lack robustness since only one or two heuristics \revise{cannot} fit all different testcases in macro legalization}.

\begin{figure}[tb!]
    \centering
    \includegraphics[width=.85\linewidth]{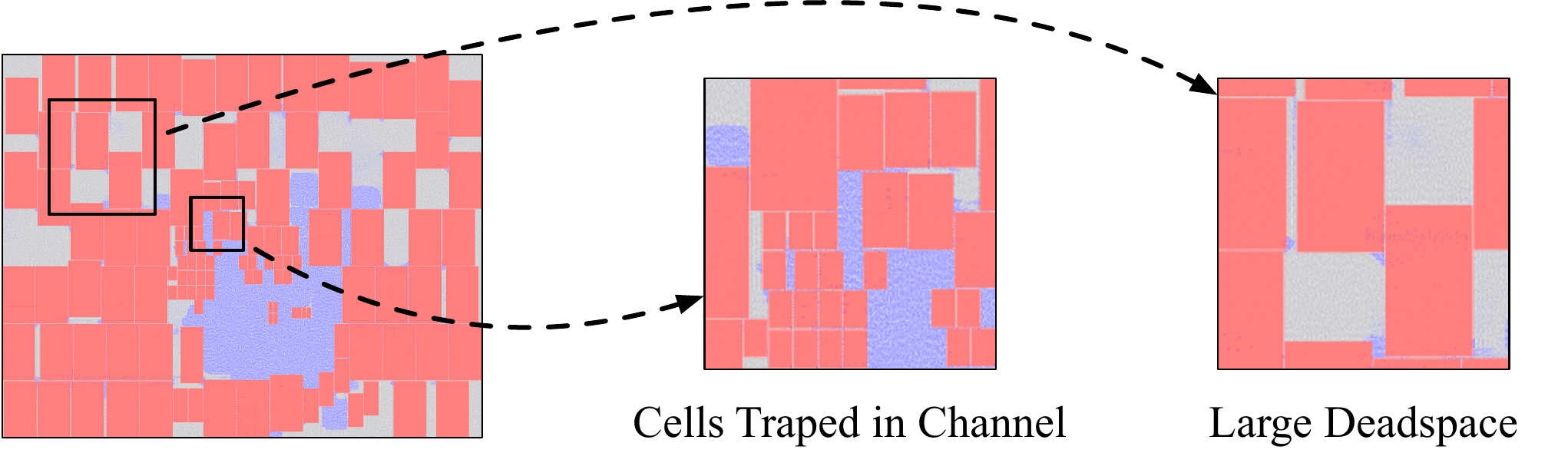}
    \caption{Consequences of ignoring regularity in macro legalization (red giant rectangles are macros, blue tiny points are cells). Both situations worsen the wirelength.}
    \label{figs:regularity-impact}
\end{figure}

\minisection{Regularity-Aware Macro Placement}
Among prior works on macro placement, there is a common focus on macro regularity. \textit{Studies have confirmed that such regularity reduces wirelength and minimizes deadspace~\cite{lin2018regularity}, although the definition of regularity in these studies is quite different.}
\cite{chang2017novel} uses cost models for macro grouping and regularity penalties, demonstrating a significant reduction in wirelength. \cite{kahng2022rtl,kahng2023hier} exploit RTL hierarchy information to cluster macros and arrange them in regular patterns. \cite{pu2024incremacro} defined macro regularity as the necessity for macros to be surrounded by other macros or die boundaries, identify irregular placements and adjust. 
The MaskRegulate~\cite{xuereinforcement} method implements a reinforcement learning policy as a regulator to adjust existing layouts. 
\cite{lin2018regularity} extract macros of the same type and similar levels, arrange them in array form. They consider regularity during refinement or \revise{directly place} macros. \textbf{However, no academic work directly considers regularity in the final legalization stage in macro placement.} Only industrial tool Innovus provides an option \texttt{place\_global\_align\_macro} to achieve regular macro legalization~\cite{innovus_manual}. 
It is important to differentiate our legalization-focused approach from constructive regular placement methods~\cite{chang2017novel,lin2018regularity,kahng2022rtl,kahng2023hier,xuereinforcement}. While constructive methods enforce patterns early, modern mixed-size global placement tends to disrupt these pre-defined patterns during optimization, so they need to keep the macros fixed. Therefore, we do not aim to compete with placement algorithms in generating the initial structure; rather, we propose an essential complementary capability: recovering and enforcing regularity during the final legalization stage, similar to the \texttt{place\_global\_align\_macro} flow in Innovus~\cite{innovus_manual}.

\begin{figure}[tb!]
    \centering
    \includegraphics[width=\linewidth]{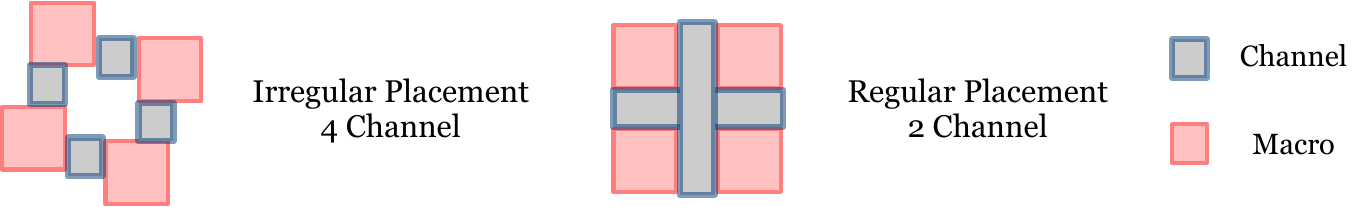}
    \caption{Channel-based macro regularity examples.}
    \label{figs:regularity}
\end{figure}

\minisection{Regularity Metrics}
\textbf{Although regularity is essential for macro placement, there is no universally accepted definition.}
Earlier studies by~\cite{lin2018regularity,kahng2022rtl,kahng2023hier} mainly characterize regularity via array-based placement configurations.  \cite{chang2017novel,pu2024incremacro,xuereinforcement} define regularity by the proximity of macros to chip boundaries or the presence of other macros in specific vicinities.

We generally adhere to the perspective in~\cite{kahng2023hier} that \textbf{mandating macro placement at the layout periphery is no longer feasible, as this increases the required stack depth of macros.} However, in~\cite{lin2018regularity,kahng2022rtl,kahng2023hier}, they constrain macros to a strictly array-based form, with no metric to evaluate their regularity.

We introduce a quantitative macro regularity metric by counting channels between macros, aligned with previous studies~\cite{lin2018regularity,kahng2022rtl,kahng2023hier}. This metric reflects these physical design principles: excessive channels increase deadspace and restrict standard-cell placement optimization. As shown in \Cref{figs:regularity-impact}, the cells trapped in the channels between macros are less likely to jump out of the channels, due to the density constraint~\cite{pu2024incremacro}. It is also evident that the deadspaces between macros increase the wirelength.
Our approach assigns peak regularity scores to squares and high scores to rectangles, supporting established array-based placement methods~\cite{lin2018regularity,kahng2022rtl,kahng2023hier}. \Cref{figs:regularity} indicates the number of channels under irregular and regular placement, which reveals that the number of channels is the smallest in the case of array-based placement.
For normalization, after counting the channels, we divide the theoretical smallest possible number of channels by the acquired number. The number is less than 1, where the closer the index value is to 1, the higher the regularity achieved.
We achieve a theoretical minimum number of channels when arranging macros in a square array. The following formula can obtain a rough estimation:
\begin{equation}
    a = \lfloor \sqrt{N_\text{macro}} \rfloor,\quad
C_{\min} = 2a - 2,
    \label{equ:channel_estimation}
\end{equation}
where $N_\text{macro}$ is the number of macros, $a$ is the side length lower bound of the square array, and $C_{\min}$ is the theoretical minimum number of channels. Intuitively, we need at least $a-1$ channels to separate the macros into $a$  rows and another $a-1$ channels to separate the macros into $a$ columns. Any irregular placement introduces extra channels.
Finally, the regularity index is defined as:
\begin{equation}
    R = \frac{C_{\min}}{C_\text{actual}},
    \label{equ:regularity_index}
\end{equation}
where $C_\text{actual}$ is the actual number of channels.
\revise{To compute $C_{\text{actual}}$, we construct a Hanan grid from all macro vertex coordinates, mark occupied cells, and iteratively merge adjacent empty cells into maximal rectangular strips. An empty cell at a crossing point may participate in merges in both directions simultaneously. The number of resulting strips yields $C_{\text{actual}}$; details are provided in \Cref{sec:prelim}.}
\Cref{figs:regularity} also exhibits that the theoretical minimum number of channels is 2. The regularity indices are 0.5 and 1 for the left and right cases, respectively.

\minisection{Large Language Models for Algorithm Design}
Large Language Models (LLMs) have drawn much research attention for their great performance in various cognitive tasks~\cite{achiam2023gpt,guo2025deepseek,wang2023scientific,ahn2024large,hongmetagpt,romera2024mathematical}. In Electronic Design Automation (EDA), more researchers are using LLMs as research tools\revise{~\cite{liu2023verilogeval,chang2024data,fu2023gpt4aigchip}}, in script/RTL generation, circuit design, and QA systems. 

\revise{
Traditional automated algorithm design methods, such as genetic programming (GP)~\cite{banzhaf1998genetic}, require defining a set of allowed mutation operations (or primitives). Designing such a suitable set of primitives is non-trivial and remains an open challenge in practice~\cite{o2010open}.
Reinforcement learning (RL) has also been applied to algorithm design, including program superoptimization~\cite{sypula2022learning} and discovering faster assembly-level sorting algorithms~\cite{mankowitz2023faster}. However, RL-based approaches require human experts to carefully design the action spaces, and this design becomes difficult for higher-level programming languages~\cite{romera2024mathematical}. Moreover, RL models are trained on specific tasks and do not readily generalize to other problems.
In contrast, LLMs have been trained on vast amounts of code and have learned common patterns and routines from human-designed programs. By leveraging this knowledge together with prompt-provided context, LLMs can generate more effective suggestions than the random mutations typically used in GP~\cite{romera2024mathematical}. Furthermore, LLMs do not require a predefined action space or mutation operations, offering greater flexibility to explore a larger search space.
}

\revise{Recently, there is growing interest in LLM-based algorithm design}~\cite{liuevolution,liu2024systematic,yanglarge,jawahar2024llm}. LLMs play various roles, such as evaluators and optimization engines. The most promising is using LLMs to directly design algorithm components (heuristic methods) for NP-hard problems\revise{~\cite{liuevolution}}. In EDA, the direct use of LLMs for algorithm design is underexplored; only~\cite{yao2025llm4placement} has attempted to enhance global placement algorithms via iterative LLM implementation, using crafted prompts and an LLM-based genetic process, but lacking meaningful feedback (only wirelength).
\revise{
AlphaEvolve~\cite{novikov2025alphaevolve} combines LLMs with evolutionary search and achieves impressive results in mathematics and algorithm optimization, where LLMs already possess sufficient knowledge from pretraining. However, applying LLMs to specialized fields like EDA, where they lack domain expertise, presents a distinct challenge.
SATLUTION~\cite{yu2025autonomous} targets NP-complete SAT problems and demonstrates that LLMs can handle substantial code generation at scale. Nevertheless, as the authors discuss in the appendix, human guidance is still incorporated to manually direct higher-level algorithmic strategies while leaving the lower-level implementation to the LLM agent.
GPU Kernel Scientist~\cite{andrews2025gpu} tunes existing GPU optimization techniques (e.g., increasing thread block occupancy and resolving shared memory conflicts) to find the best kernel for specific hardware, rather than discovering new algorithms.
Our work differs from these approaches in three key aspects:
(1)~we introduce a \textbf{domain-agnostic abstraction} that reformulates the EDA-specific problem into a geometric problem. Domain-specific decisions such as clustering strategy and optimization objective selection are encoded in the framework by human engineers, while the LLM operates solely on the abstracted geometric subproblem using a general-purpose, off-the-shelf model without any domain-specific fine-tuning.
(2)~Our LLM autonomously discovers both algorithm ideas and implementations \textbf{without human strategy guidance} in the intermediate process.
(3)~The manageable scale of macro legalization instances enables us to provide human-crafted solution samples and per-iteration visual feedback to the LLM, allowing it to understand \textbf{why} a solution is good or bad rather than relying solely on scalar metrics.
}

Despite their promising potential, significant limitations persist in employing LLMs for advanced algorithm design: they struggle with complex algorithm implementations and lack domain-specific expertise for specialized optimization tasks~\cite{liu2024systematic}.

\minisection{Motivation}
In the EDA domain, the majority of problems are NP-hard. Researchers have historically developed numerous heuristic algorithms to address these challenges. Specifically within macro legalization, previous researchers have designed sophisticated heuristics for macro clustering, placement selection, and perturbation techniques to optimize solution quality. It naturally raises a fundamental question: Can LLMs effectively replace human experts in designing heuristic algorithms for such problems? 

\minisection{Our Contribution}
To address these challenges, we propose a robust regularity-aware macro legalization framework with four stages.
Our framework decomposes the macro legalization process into four sequential components of
clustering, regularity-aware contour generation, template-based macro matching, and inter-cluster refinement as shown in~\Cref{figs:introduction}.
\textbf{Compared with the traditional legalization methods, our methods are more robust through the combination of multiple heuristic LLM-designed contour algorithms.
It achieves simultaneous regular macro legalization while minimizing the displacement through template matching.}
To mitigate the limitations of LLMs in EDA domains, we extract a domain-agnostic geometry problem and present LLMs with the engineers' manual macro legalization solutions, which address gaps in domain knowledge and simplify the original problem. This enables fast, automatic, and diverse heuristics findings in an offline manner to fit different input testcases (no human intervention in the intermediate process). Notably, the hard testcases used for final validation are distinct from those utilized to guide LLMs in heuristic discovery, effectively demonstrating the strong generalization capability of LLM generated heuristics.
\revise{We validate MacroAgent under two experimental flows. In the academic flow (DREAMPlace + HeLEM-GR), MacroAgent reduces routed wirelength by 3\%--5\% with comparable congestion on both TILOS and Chipyard benchmarks. In the industrial flow (Cadence Innovus place-and-route), MacroAgent achieves 2.9\% lower routed wirelength, 68.3\% TNS improvement, and comparable power over the DREAMPlace macro legalization baseline; it also achieves 1.8\% lower routed wirelength when integrated into the Innovus macro placement flow.}

We organize the rest of the paper as follows:
\Cref{sec:prelim} introduces the rise of LLM agents in algorithm design and the problem definition of macro legalization.
\Cref{sec:algo} introduces the proposed macro legalization framework.
\Cref{sec:impl} introduces the implementation details of the proposed macro legalization framework.
\Cref{sec:experiment} presents the experimental results.
\Cref{sec:conclusion} concludes the paper.

\section{Preliminaries}
\label{sec:prelim}


\subsection{LLM Agents for Algorithm Design}
\label{sec:llm_agents}
An LLM agent is an AI that uses LLMs to autonomously perform tasks by reasoning, taking step-by-step actions, and leveraging tools.
Currently, LLM agents are revolutionizing algorithm design through automated code generation and optimization.
\revise{Recent studies~\cite{liu2024systematic} identify several distinct roles that LLMs play in algorithm design:
(1)~\textbf{LLM as optimizer}: OPRO~\cite{yanglarge} uses LLMs to directly generate and refine solutions through prompt-based optimization, treating the LLM itself as the search operator;
(2)~\textbf{LLM as evaluator}: Jawahar et al.~\cite{jawahar2024llm} leverage LLM performance predictors as initializers for neural architecture search, where the LLM assesses candidate quality rather than generating algorithms;
(3)~\textbf{LLM as heuristic generator}: this is the most promising paradigm for NP-hard problems, where LLMs generate complete algorithm code within an evolutionary framework.
}
Google's FunSearch~\cite{romera2024mathematical} and AlphaEvolve~\cite{novikov2025alphaevolve} \revise{exemplify this third paradigm},
demonstrating AI's capacity to discover and refine algorithms across scientific and engineering domains.
\revise{EoH~\cite{liuevolution} further advances this direction by evolving both algorithmic ideas and their implementations simultaneously.}
The core idea is evolutionary code generation and evaluation. LLM agents autonomously generate, modify, and evolve complete algorithm code over multiple iterations. And evaluation systems guide the algorithmic evolution process toward better solutions. This methodology essentially creates a self-improving cycle to continuously refine results.
\revise{
    Traditional automated algorithm design methods rely on genetic programming or reinforcement learning, both of which require experts to design the mutation operations or action spaces.
    In contrast, LLM agents directly generate algorithms within the algorithm space using high-level programming languages.
    In EDA, EvoPlace~\cite{yao2025llm4placement} applies this paradigm to global placement by iteratively evolving placement heuristics via LLM-based code generation.
}

\subsection{Macro Legalization Problem Formulation}
\begin{myproblem}
    Given macro \revise{positions} after mixed-size placement  $M = \{m_{1}, m_{2}, \cdots, m_{n}\}$, where each macro $m_{i}$ has (1) initial coordinates $(x_{i}^{\prime}, y_{i}^{\prime})$ (representing bottom-left corner), (2) fixed dimensions: width $w_{i}$ and height $h_{i}$.
    Find legal macro positions $\{(x_{1},y_{1}),(x_{2},y_{2}),\cdots, (x_{n},y_{n})\}$ 
that:
\begin{enumerate}
    \item \textbf{No Overlap}: For any two distinct macros $m_{i}, m_{j}\in M$, there is no overlap between them, as depicted in~\Cref{equ:no_overlap}.
    \begin{equation}
         \left[ x_{i}, x_{i} + w_{i} \right] \times \left[ y_{i}, y_{i} + h_{i} \right] \cap \left[ x_{j}, x_{j} + w_{j} \right] \times \left[ y_{j}, y_{j} + h_{j} \right] = \varnothing, 
        \label{equ:no_overlap}
    \end{equation}

    \item \textbf{Minimum Displacement}: Minimize the total Manhattan displacement of all macros. The displacement of $m_{i}$ is $|x_{i}' - x_{i}| + |y_{i}' - y_{i}|$, so the displacement objective is 
    \begin{equation}
       \min \sum_{i=1}^n \left(|x_{i}' - x_{i}| + |y_{i}' - y_{i}|\right),
       \label{equ:minimum_displacement}
    \end{equation}

    \item \textbf{Maximum Regularity}: Regularity index $R$ is quantified by the channel ratio metric.
    For a macro placement, $R$ is defined as:
    \begin{equation}
        R = \frac{C_{\text{min}}}{C_{\text{actual}}},
        \label{equ:maximum_regularity}
    \end{equation}
    where $C_{\text{min}}$ \revise{refers to the theoretical minimum number of channels} and $C_{\text{actual}}$ is \revise{the actual number of channels}.

    \revise{\textbf{Channel Counting Method:} To compute $C_{\text{actual}}$ in practice, we employ a Hanan-grid-based procedure. First, we collect all $x$- and $y$-coordinates of every macro vertex and construct a Hanan grid, i.e., a rectilinear grid whose lines pass through every macro boundary. Each grid cell occupied by a macro is marked as \textbf{occupied}; the remaining cells are \textbf{empty}. We then iteratively merge adjacent empty cells into maximal rectangular strips: vertically adjacent empty cells sharing the same width are merged, and horizontally adjacent empty cells sharing the same height are merged. An empty cell at a channel crossing point may participate in merges in both directions simultaneously (i.e., it is shared by multiple strips rather than consumed by one). The merging repeats until no further merges are possible, and the number of resulting merged strips equals $C_{\text{actual}}$. This method naturally handles general layouts: more irregular placements produce more fragmented strips, yielding higher $C_{\text{actual}}$ and thus lower regularity index $R$.}
\end{enumerate}
\end{myproblem}

\section{Algorithm Design}
\label{sec:algo}

\subsection{Overview of MacroAgent}

Our proposed macro legalization consists of four principal components: size-aware distance-based clustering, regularity-aware contour generation, template-based macro matching, and inter-cluster refinement.
We identify regularity-aware contour generation as the most influential component, as it directly enables macros within a cluster to be arranged in a highly regular configuration. 
We employ an LLM agent to optimize this critical component.
\revise{This choice follows the LLM-based algorithm discovery paradigm~\cite{romera2024mathematical,novikov2025alphaevolve}, which requires an automated evaluator for the iterative generate-evaluate loop.
Matching already admits an optimal solution via the Hungarian algorithm, leaving no room for improvement.
Clustering lacks a direct quality metric: its output can only be assessed after running the downstream pipeline, whose key component (contour generation) is itself heuristic. This makes the end-to-end result quality unreliable and creates a credit assignment problem that hinders LLM optimization.
In contrast, contour generation's immediate downstream is matching, which is optimally solved by the Hungarian algorithm; therefore, the result quality of a contour algorithm can be evaluated directly and reliably via regularity and displacement metrics.}

\minisection{Overview} \textit{The cluster step determines which macros are suitable for regular placement. The use of regularity-aware contour generation helps achieve  regularity and diverse regular patterns for robustness; template-based macro matching can minimize displacements; and inter-cluster refinement further ensures no overlap — all of these techniques working together to generate high-quality macro legalization results.}

\subsection{Size-Aware Distance-Based Clustering}
\label{sec:clustering}
Consistent with prior research, we first cluster macros. 
Previous approaches usually determine clustering by macro dimensions, interconnection relationships, positions in the placement prototype, 
or hierarchical associations~\cite{chang2017novel,lin2018regularity,kahng2022rtl,kahng2023hier}. 
Our clustering method simplifies this by considering only macro dimensions and positions in the placement prototype, as interconnection relationships were partly accounted for in the mixed-size global placement stage. 
Since many similar heuristic algorithms have been proposed before, we utilize established methods instead of LLM-based approaches for this algorithm design.

\begin{algorithm}
    \caption{Size-Aware Distance-Based Macro Clustering.}
    \label{alg:macro_clustering}
    \begin{algorithmic}[1]
    \Require Macro set $\mathcal{M}$, size grouping factor $\alpha$, proximity factor $\beta$
    \Ensure Assignment of macros to clusters
    \State $clusterID \gets 0$;
    \State Size grouping: $groups \gets$ group macros by size using $\alpha$;
    \For{each size group $g \in groups$}

            \State construct KD-tree from macro center points;
            \For{each unvisited macro $m \in g$}
                \State $current \gets clusterID$ then increment $clusterID$; 
                \State BFS traversal of neighboring macros;
                \State \ \ For current macro $m_c$, calculate  $T_x = \revise{w}(m_c) \cdot \beta$, $T_y = \revise{h}(m_c) \cdot \beta$;
                \State \ \ Find all macros within \revise{horizontal} distance $T_x$ and \revise{vertical} distance $T_y$;
                \State \ \ Assign them to $current$ and continue traversal;
            \EndFor

    \EndFor
    \end{algorithmic}
\end{algorithm}

\Cref{alg:macro_clustering} comprises two procedures: dimensional classification and spatial clustering.
The algorithm first utilizes parameter $\alpha$ to partition macros based on dimensional similarity. Dividing the macro width and height by $\alpha$, yields dimension identifiers, which are then used to group macros with close dimensions.
Subsequently, within each dimensional group, the algorithm employs parameter $\beta$ in conjunction with a KD-tree to perform proximity-based clustering.
In our experiments, we fix $\alpha = 10$ and $\beta = 1.5$. Specifically, $\alpha = 10$ means that we treat macros' size difference less than 10 site width as the same size (as DREAMPlace scales site width to 1). $\beta = 1.5$ means that we consider macros with a distance less than 1.5 times their respective sizes as neighboring macros. In pilot trials on the easy cases, we observed that modest variations around these values only lead to slight changes in cluster granularity.
Notably, we implement \revise{adaptive} proximity thresholds ($T_{x}$ and $T_{y}$) that are proportional to macro's dimension sizes, enabling larger macros to have expanded search domains. In comparison, smaller macros maintain narrower search scopes.
This clustering efficiently aggregates macros with similar dimensions and spatial proximity into clusters.
\revise{Note that $\alpha$ serves solely as a grouping granularity for this clustering stage; once clusters are formed, all subsequent stages---contour generation (\Cref{sec:llm}), template generation, and macro matching---operate on the original, unscaled macro dimensions. Likewise, the regularity metric $R$ in \Cref{equ:maximum_regularity} and channel counting are evaluated using the actual macro sizes and legalized positions.}

\subsection{Regularity-Aware Contour Generation}
\begin{figure}[tb]
    \centering
    \begingroup
    \setlength{\tabcolsep}{1.5pt}
    \begin{tabular}{cccc}
    \subfloat[Initial ]{\includegraphics[width=0.24\textwidth]{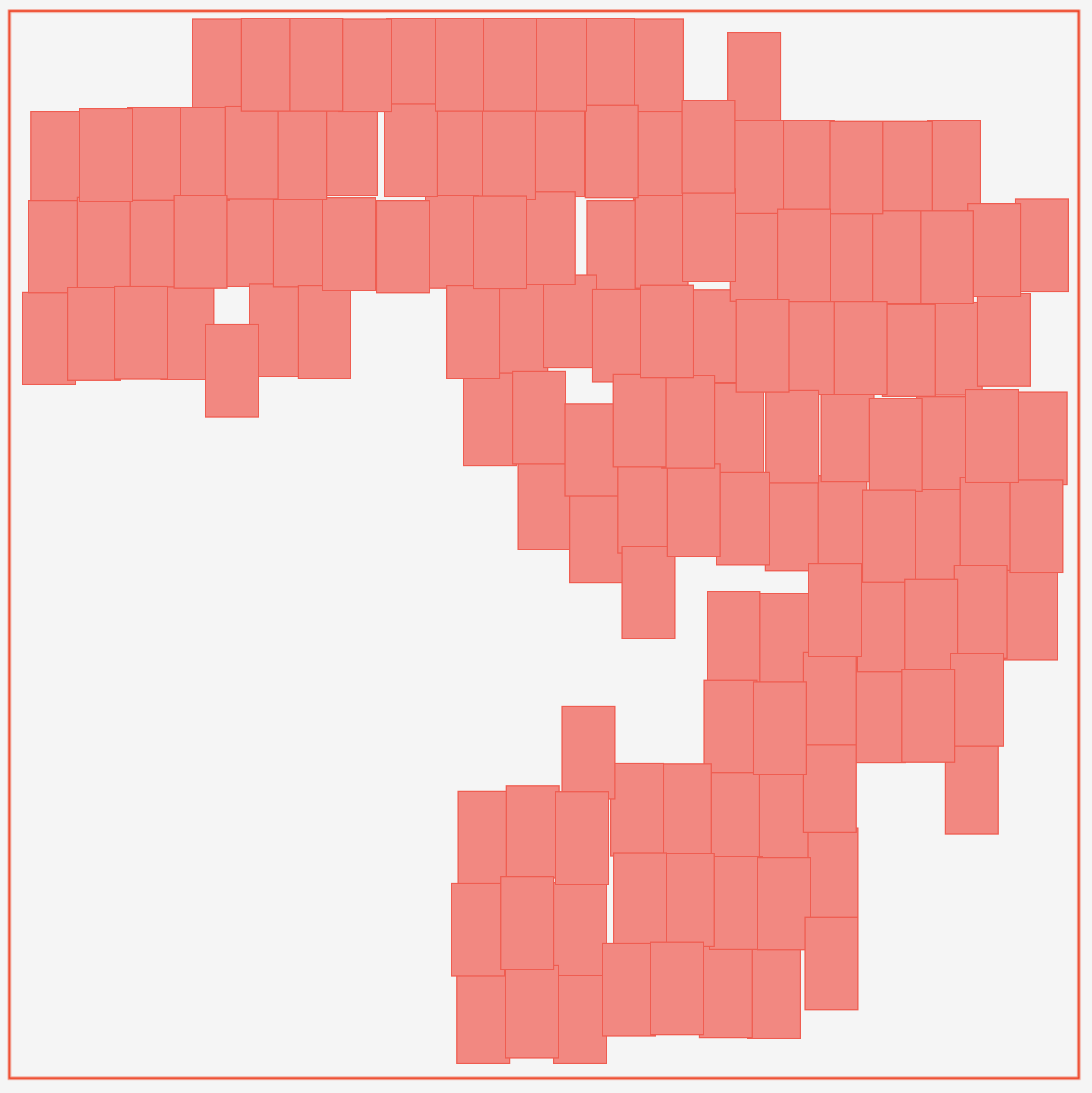}}&
    \subfloat[Grid ]{\includegraphics[width=0.24\textwidth]{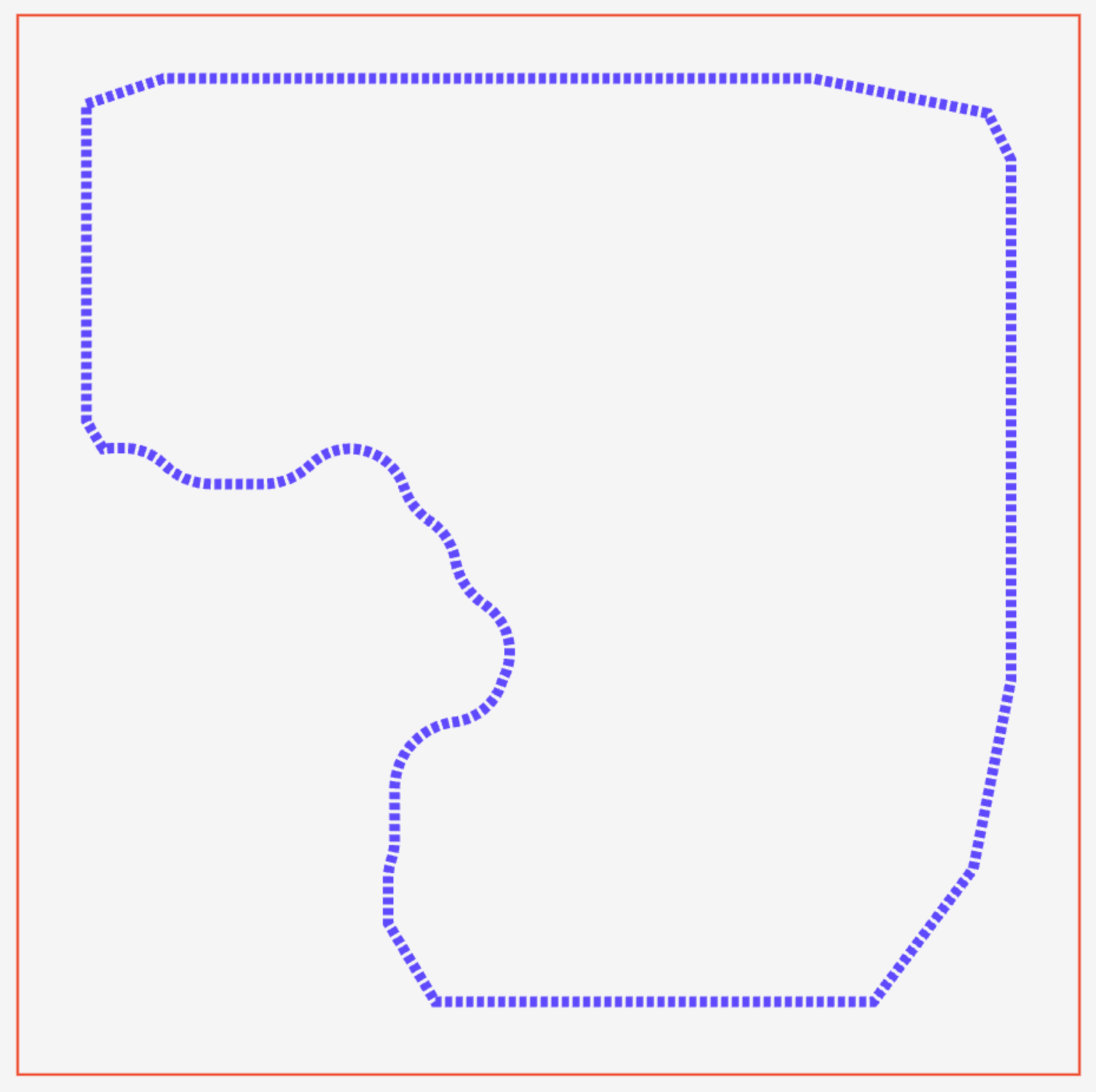}}&
    \subfloat[Alpha shape ]{\includegraphics[width=0.24\textwidth]{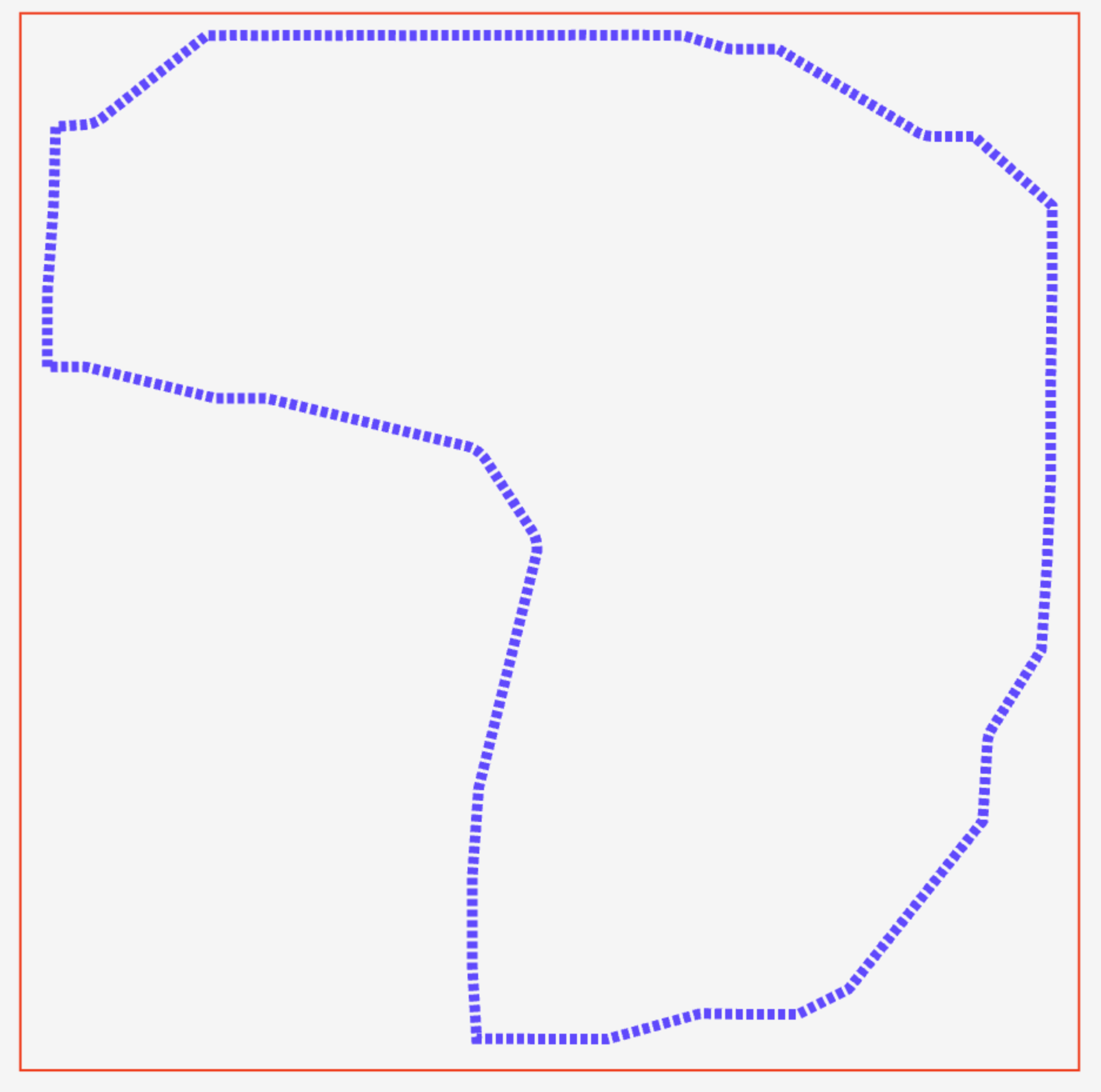}}&
    \subfloat[MST ]{\includegraphics[width=0.24\textwidth]{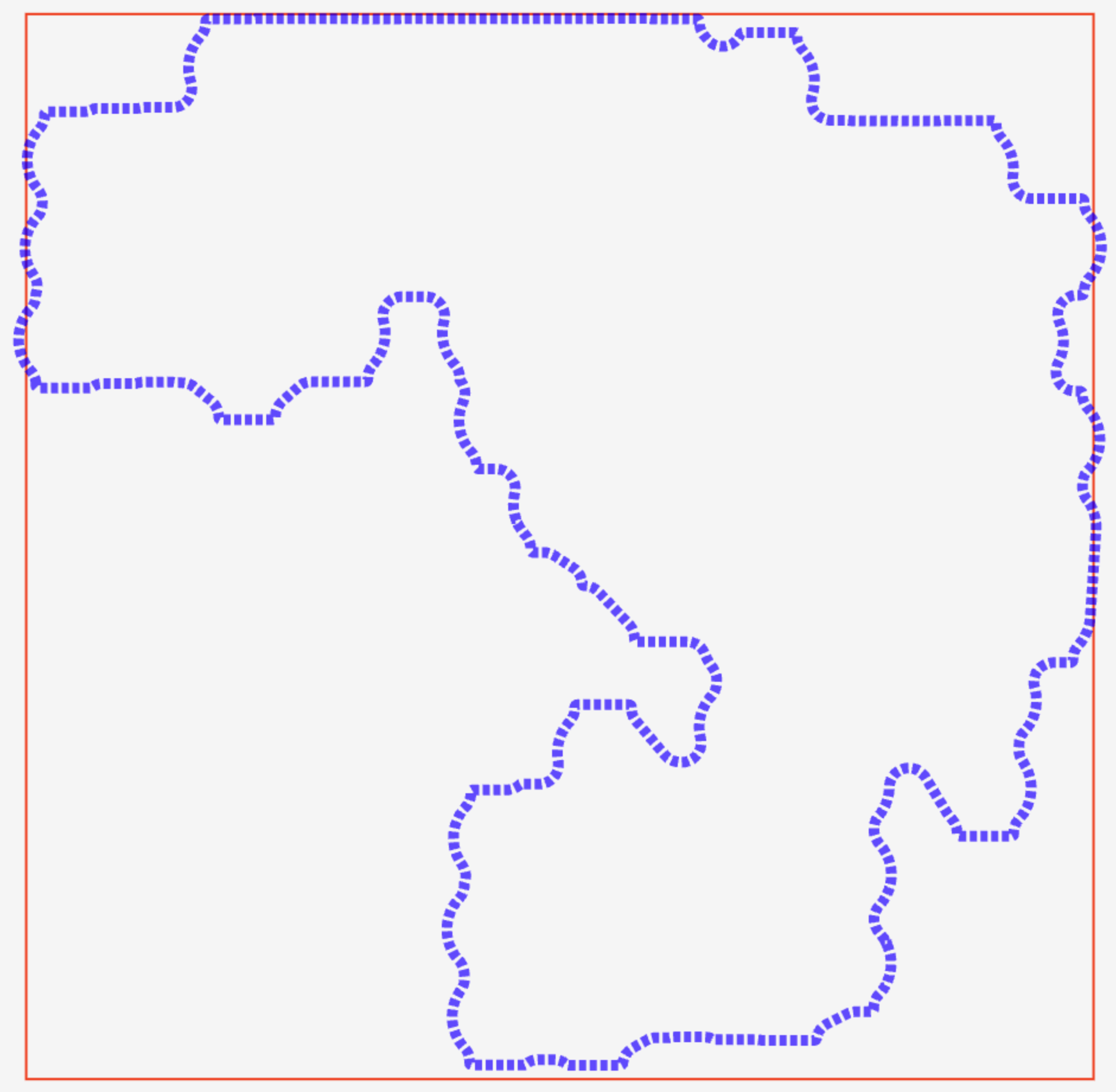}}\\[4\jot]
    \subfloat[Initial ]{\includegraphics[width=0.24\textwidth]{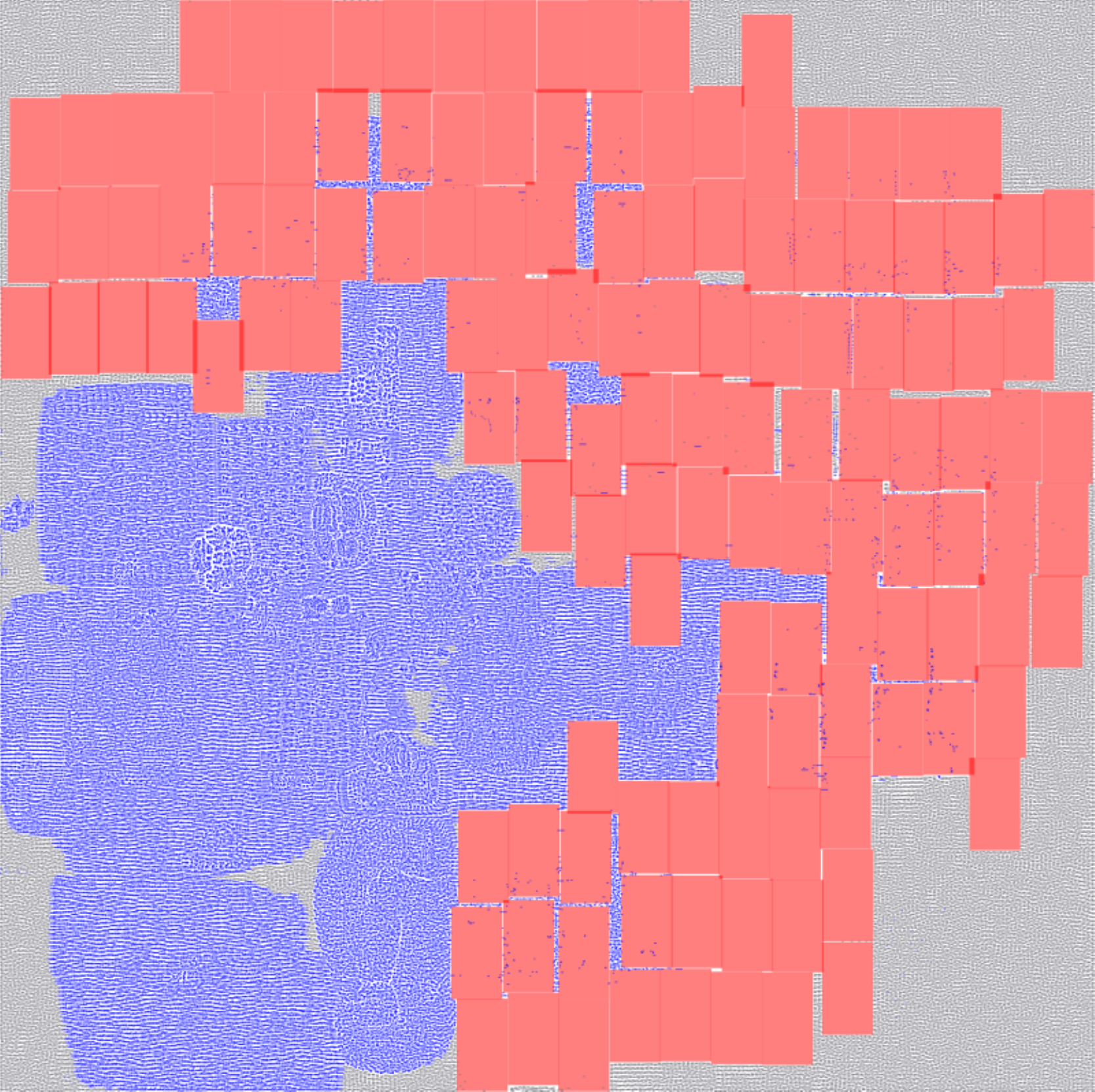}}&
    \subfloat[Grid ]{\includegraphics[width=0.24\textwidth]{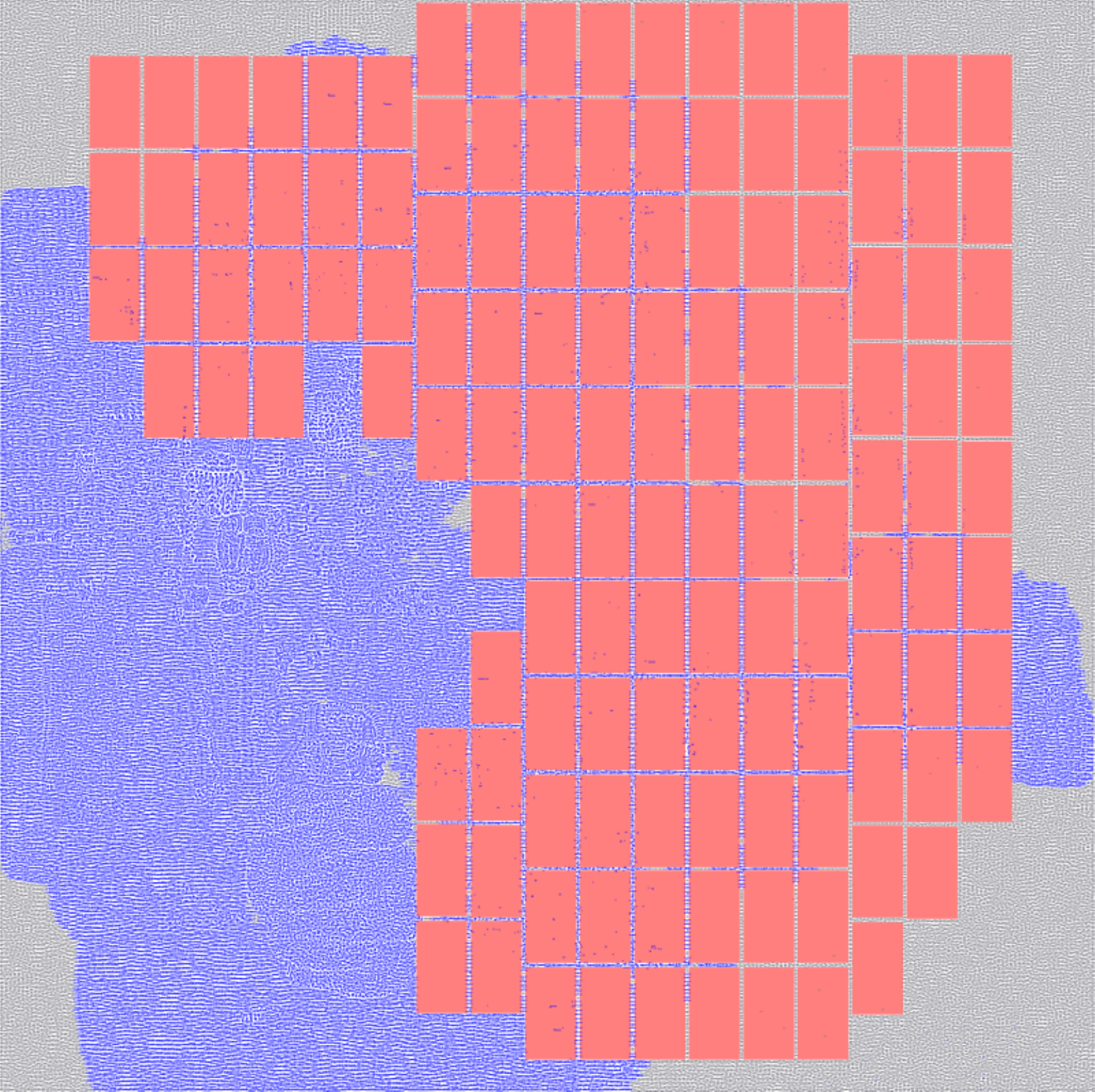}}&
    \subfloat[Alpha shape ]{\includegraphics[width=0.24\textwidth]{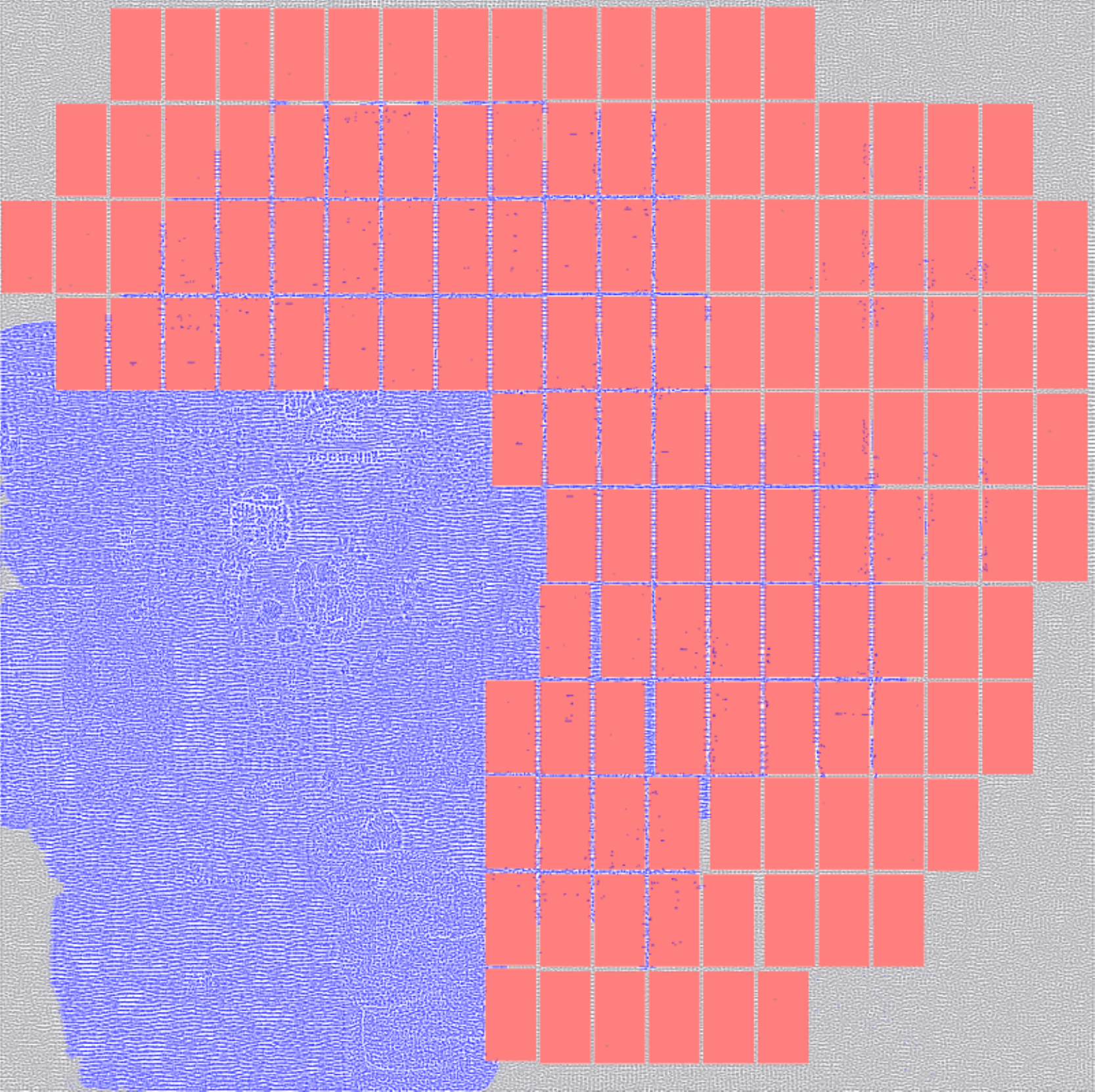}}&
    \subfloat[MST ]{\includegraphics[width=0.24\textwidth]{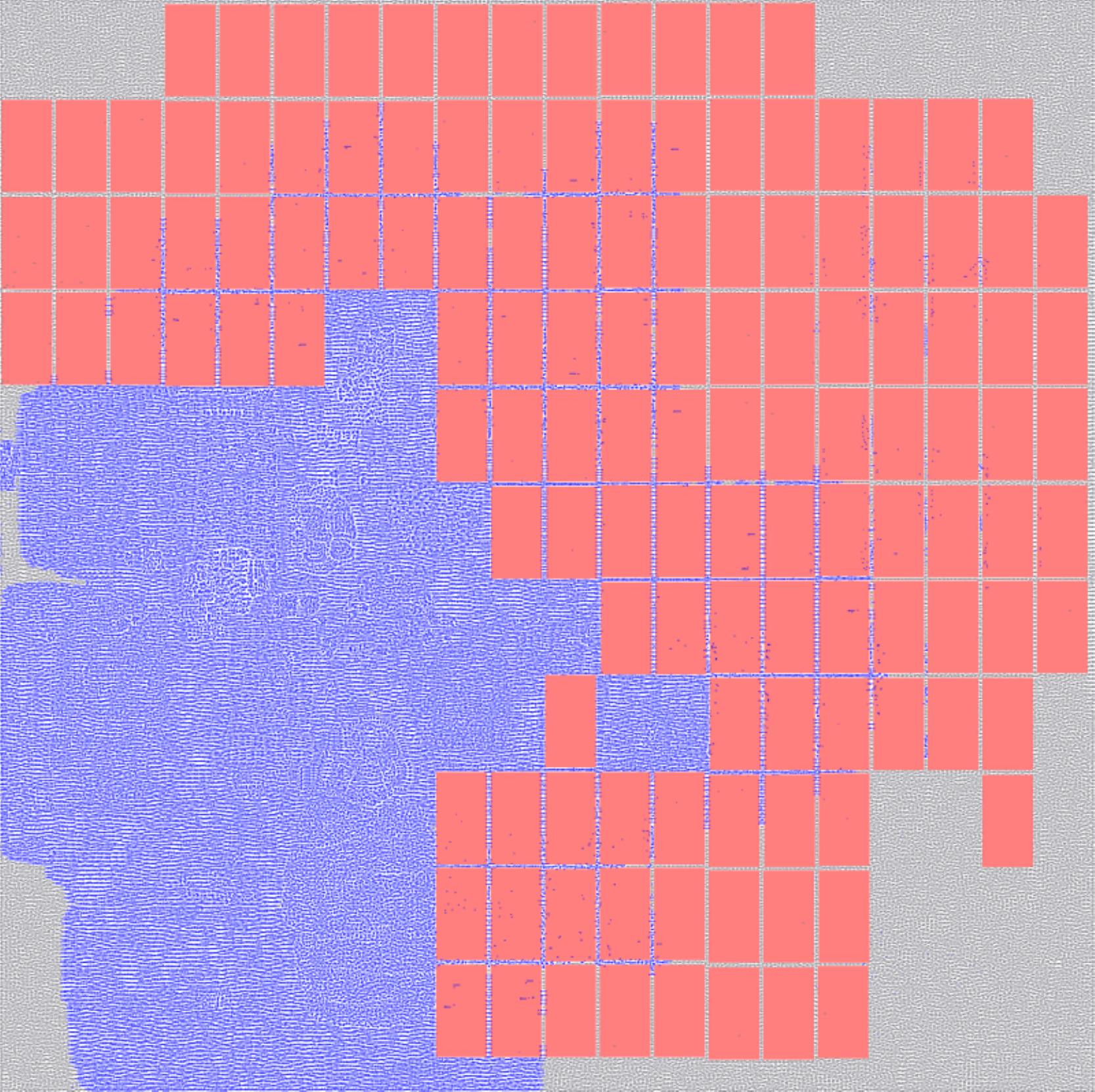}}
    \end{tabular}
    \endgroup
    \caption{Visualizations of different contour algorithms in \texttt{Ariane133}: (top row) contour visualization and (bottom row) layout after standard-cell placement (red for macros; blue for standard cells). From left to right, regularity \revise{decreases}, but displacement \revise{improves}. Initial is for original global placement outputs, and Grid, Alpha shape, and MST are for different legalization results. }
    \label{fig:viz_combined}
\end{figure}

Regularity-aware contour generation is the key step to achieve high regularity while minimizing displacement.
Previous research tends to place the clustered macros in a rectangular formation~\cite{lin2018regularity,kahng2022rtl,kahng2023hier}. In other words, 
the contour of the macro cluster is a rectangle, as it is the simplest regular shape. However, our empirical observations reveal that many macro clusters after mixed-size placement exhibit contours that deviate significantly from rectangular shapes as shown in top-left figure in \Cref{fig:viz_combined}. 
Enforcing rectangular arrangements of macro clusters often results in displacement-induced degradation of design quality.
Consequently, we propose generating diverse regularity-aware contours based on the outline morphology of macro clusters after mixed-size placement.
The regularity part of contour algorithms lies in the removal of some burrs on the contour, as shown in \Cref{fig:viz_combined}. The fewer burrs, the better the regularity, but the worse the displacement.
\textbf{We compare the manual designed rectangle contour \revise{heuristic} with LLM-designed contour generation \revise{heuristics} in the experiments \Cref{fig:hpwl-ratio-avg-line} to prove that LLM-designed methods can achieve better design quality.}

\minisection{Necessity of Multiple Contours}.
The reason for diverse contours is that regularity and displacement are conflicting, so we need to trade them off; furthermore, there are no direct analytical solutions to guide this trade-off (only evident after standard-cell placement).
We here present the results of \texttt{Ariane133} under different contours. \Cref{fig:viz_combined} shows the three macro legalization results generated for \texttt{Ariane133}. The picture on the far left is the result of an initial mixed-size placement. The three pictures on the right are the results generated by three different contour algorithms, respectively.
Smoothing contour burrs improves the channel metric (higher regularity) but typically increases macro displacement; different contours (Grid, Alpha‑shape, MST) realize different points on this trade‑off frontier.

\begin{figure*}[tb!]
    \centering
    \includegraphics[width=\linewidth]{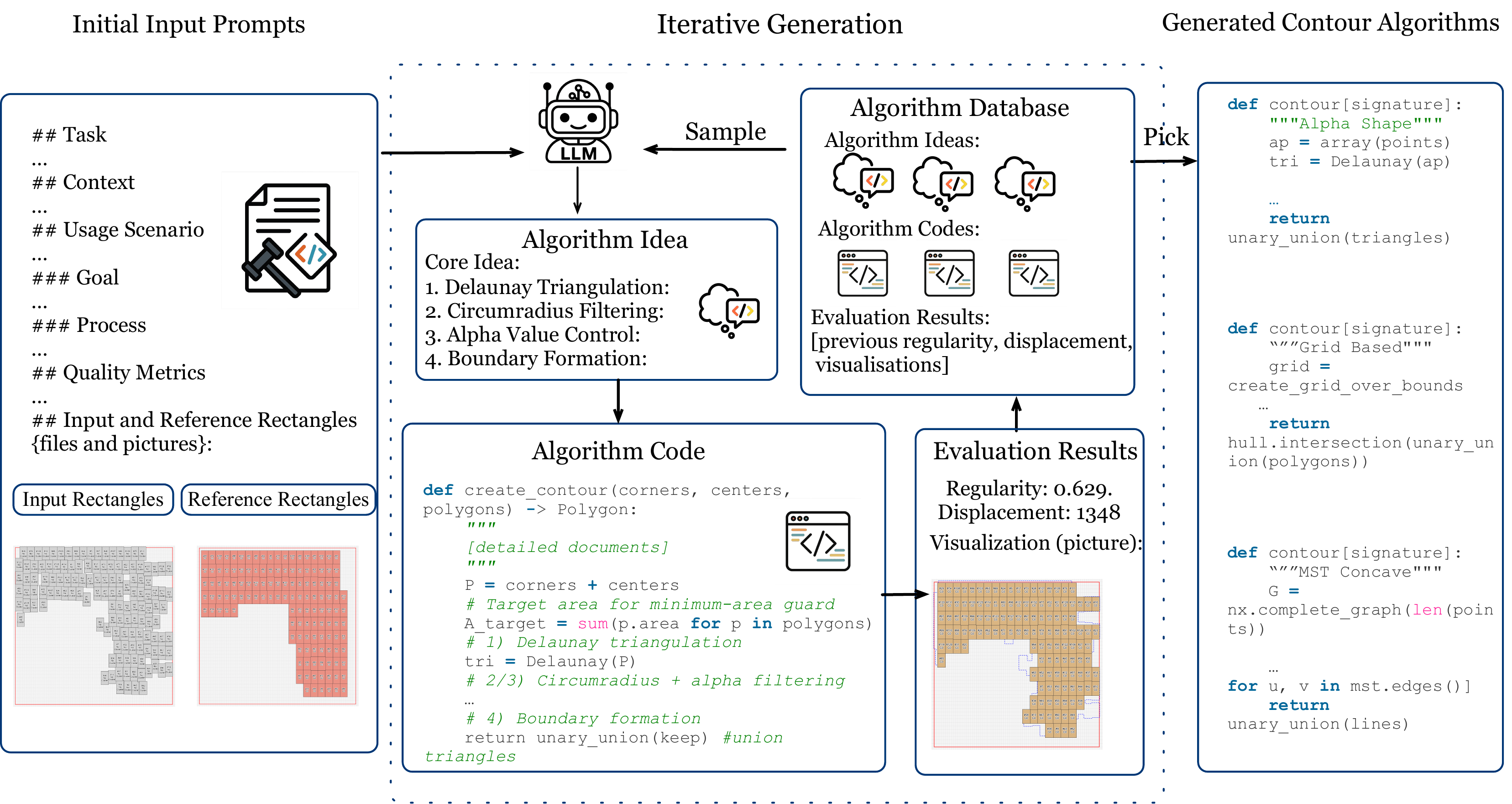}
    \caption{LLM agent-based contour algorithm generation flow. \revise{Key components: (1) \textbf{Input Prompt:} Contains task description, metrics (regularity, displacement), previous ideas/code, evolution directive (E1/E2/M1/M2/M3), and reference solutions; (2) \textbf{LLM Agent:} Processes the prompt and generates new algorithm ideas and executable Python code; (3) \textbf{Code Execution:} Runs the generated contour algorithm on test cases; (4) \textbf{Evaluation:} Computes regularity and displacement metrics, generates visualizations; (5) \textbf{Algorithm Database:} Stores successful algorithms that outperform existing ones on any test case; (6) \textbf{Feedback Loop:} Results feed back into the next iteration's prompt to guide further evolution.}}
    \label{figs:agent}
\end{figure*}

\begin{figure*}[tb!]
    \centering
    \includegraphics[width=0.8\linewidth]{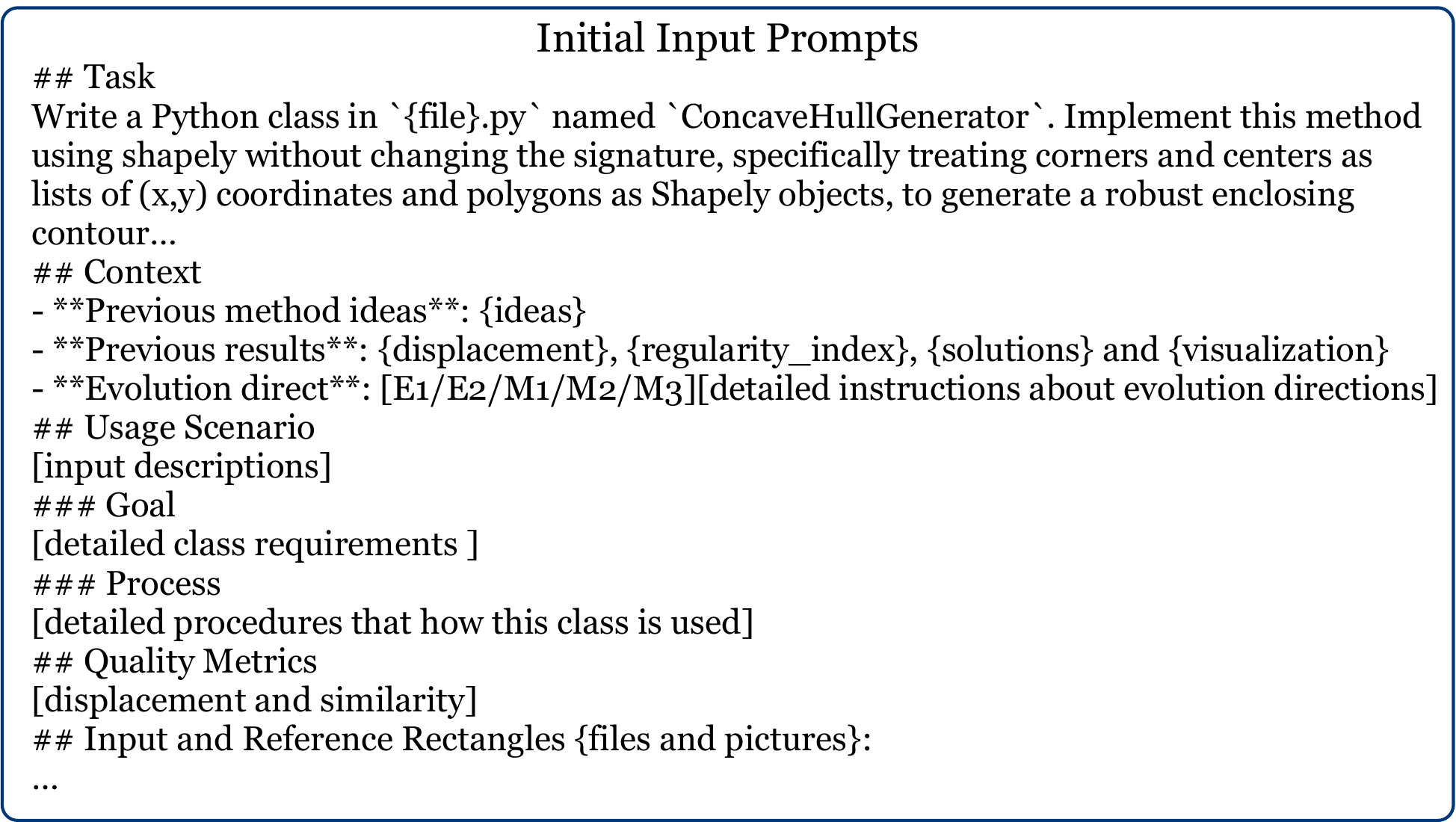}
    \caption{Simplified version of the ``initial input prompt'' for LLM agent-based contour algorithm generation. \revise{Main sections: (1) \textbf{Task Description:} Defines the contour generation problem and its role in macro legalization; (2) \textbf{Metrics:} Specifies regularity (channel count) and displacement objectives; (3) \textbf{Previous Ideas:} Provides natural-language descriptions of prior algorithms (e.g., rectangle, convex hull); (4) \textbf{Previous Code:} Shows executable Python implementations of prior algorithms; (5) \textbf{Evolution Strategy:} Instructs the LLM on how to modify/improve (E1: explore diversity, E2: synthesize and innovate, M1: fix weaknesses, M2: tune parameters, M3: simplify); (6) \textbf{Reference Solutions:} Includes manual legalization examples to guide the LLM toward practical solutions.}}
    \label{figs:prompt}
\end{figure*}

\begin{figure}[tb!]
    \centering
    \includegraphics[width=\linewidth]{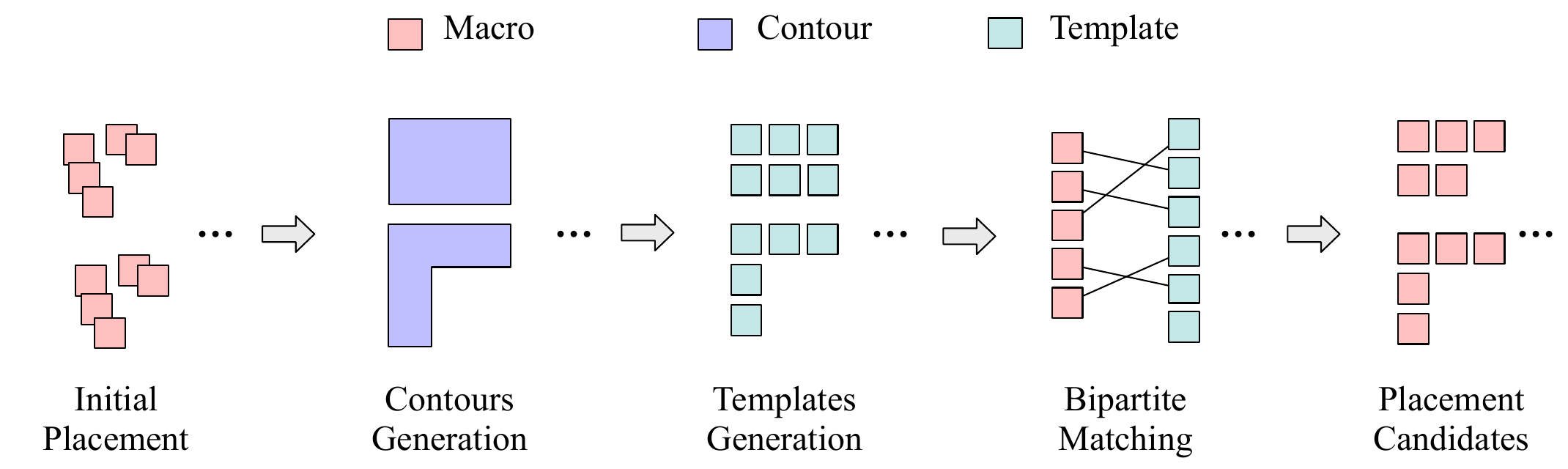}
    \caption{\revise{Template-based macro matching. The process consists of three stages: (1) a regularity-aware contour is generated around the macro cluster; (2) a uniform grid template is tiled inside the contour and feasible grid cells are selected; (3) macros are assigned to grid positions via the Hungarian algorithm to minimize total displacement.}}
    \label{figs:template_matching}
\end{figure}
\minisection{LLM Agent-Based Contour Algorithm Generation Flow}\label{sec:llm}
\Cref{figs:agent} demonstrates the end-to-end workflow for discovering contour-generation algorithms with the proposed LLM agent. Prompts and code are simplified due to page size limits.
Initially, we craft a comprehensive prompt that specifies (1) task, (2) usage scenario, (3) quality metrics (regularity, displacement), (4) prior ideas/code, (5) evolution directive (one of E1/E2/M1/M2/M3), (6) outcomes, and (7) reference manual legalization solutions.  The content of the simplified version of the ``initial input prompt'' is shown in \Cref{figs:prompt}.
At the first initialize round, the ``previous idea'' is the simple \textbf{rectangle contour} generation algorithm: it computes the bounding box of the macro cluster, scales the rectangle to match the cluster's area, and recenters it on the cluster.

\revise{Specifically, the workflow in~\Cref{figs:agent} comprises six key components:
(1)~\textbf{Input Prompt} encapsulates the complete problem specification fed to the LLM, including the task description, evaluation metrics, prior algorithm ideas and code, an evolution directive, and reference solutions (detailed below with \Cref{figs:prompt});
(2)~\textbf{LLM Agent} receives the assembled prompt and generates a new algorithm: a natural-language \textit{idea} describing the heuristic logic, together with an executable Python \textit{code} implementation of the contour generation function;
(3)~\textbf{Code Execution} runs the generated contour algorithm on the macro cluster testcases, producing contour polygons that are subsequently used for template generation and macro matching;
(4)~\textbf{Evaluation} computes the regularity metric $R$ and displacement for each testcase and generates layout visualizations, enabling quantitative comparison against existing algorithms;
(5)~\textbf{Algorithm Database} archives every algorithm whose regularity or displacement outperforms all existing entries on at least one testcase, forming a growing Pareto-front repository;
(6)~\textbf{Feedback Loop} feeds the evaluation results and the current algorithm database back into the next iteration's prompt, allowing the LLM to learn from prior successes and failures and progressively refine its designs.}

\revise{\Cref{figs:prompt} shows a simplified version of the input prompt, which is organized into six sections:
(1)~\textbf{Task Description} defines the contour generation problem: given a set of macro positions and dimensions within a cluster, produce a closed rectilinear polygon that encloses the macros with appropriate regularity.
(2)~\textbf{Metrics} specifies the two optimization objectives: regularity (measured by the channel count ratio $R$) and displacement (Manhattan distance from original positions), establishing the evaluation criteria the LLM must optimize.
(3)~\textbf{Previous Ideas} provides natural-language descriptions of prior contour algorithms (e.g., rectangle bounding box, convex hull, alpha shape), giving the LLM conceptual building blocks to draw upon or diverge from.
(4)~\textbf{Previous Code} includes the executable Python implementations corresponding to the previous ideas, enabling the LLM to understand concrete implementation patterns and reuse effective code structures.
(5)~\textbf{Evolution Strategy} instructs the LLM on \textit{how} to modify or improve the previous algorithms using one of five directives adopted from~\cite{liuevolution}.
Two are exploration strategies: E1 selects $p$ parent heuristics and asks the LLM to generate a maximally different new heuristic; E2 first summarizes the common ideas behind $p$ parents and then creates a new heuristic that builds on those shared insights yet diverges substantially.
Three are modification strategies: M1 selects a single parent and asks the LLM to diagnose its weaknesses and revise both idea and code; M2 adjusts only the parameters of a single parent while preserving its algorithmic structure; M3 identifies and removes redundant components to simplify the implementation.
(6)~\textbf{Reference Solutions} includes manually crafted macro legalization examples for the testcases, providing the LLM with concrete targets that demonstrate what high-quality placements look like and guiding it toward practically effective contour shapes.}

Then, we design a generation flow that evolves both \textit{ideas} and \textit{code} via the agent flow adopted from~\cite{liuevolution}. 
The process jointly optimizes \textbf{ideas} (natural-language heuristic logic) and \textbf{code} (executable Python functions) under an \textbf{initialize → generate → evaluate → iterate} framework: (1) \textbf{Initialize}: Create $N$ initial heuristics via the initial input prompt; (2) \textbf{Generate}: Until the \textbf{G-generation} stopping criterion, we apply \textbf{5 prompt strategies} to produce  new candidates per generation. Previous ideas are sampled probabilistically (sample number p=5 for E1/E2; p=1 for M1–M3). Prompts combine previous \textit{ideas} and \textit{code} to produce varied outputs across conceptual and implementation domains; (3) \textbf{Evaluate}: Evaluate on problem testcases; admit only feasible solutions to the algorithm database; (4) \textbf{Iterate}: Sample the \textbf{p} candidates from the algorithm database to form the next generation; repeat until completion, then output the best heuristics.
We propose five prompt strategies to improve the performance: (1) \textbf{E1 (exploration)}: Generate heuristics maximally different from the five samples to expand diversity; (2) \textbf{E2 (exploration with synthesis)}: First summarize common ideas across the five samples, then propose substantially different heuristics to ensure effective innovation; (3) \textbf{M1 (structural modification)}: Diagnose a single sample's shortcomings and revise both idea and code to improve performance; (4) \textbf{M2 (parametric tuning)}: Adjust parameters only, preserving the algorithmic structure; (5) \textbf{M3 (simplification)}: Remove redundant code components to improve efficiency. 

To control cost and ensure reproducibility, we adopt a fixed budget: $N=10$ initial heuristics and $G=20$ generations. These hyperparameters are aligned with the prior work~\cite{liuevolution}. The only difference is that here, to control the cost, we only use each prompt strategy once instead of $N$ times in~\cite{liuevolution}. In total, we generate $N + G \times 5 = 110$ heuristics.

We modify the population management strategy in the prior work~\cite{liuevolution}. Instead of using a fixed size population, we decide to keep the 
heuristics if they have better regularity or displacement than existing ones on any testcases.

We add extra prompts before algorithm idea/code generation. These prompts make the LLM output structured contents we need. Usually, the LLM follows the structures (function signature) mentioned in the prompts. If the LLM contradicts the structure we need, we directly discard its output and start next iteration. Since this rarely happens, this method does not hinder the process.



\subsection{Template-Based Macro Matching}
\label{sec:matching}
After generating regularity-aware contours, we create templates based on them.
Here, a template refers to a grid structure where each cell in the grid can accommodate a macro. Utilizing the template naturally leads to regular macro legalization, and our objective is to minimize the displacement. \revise{Matching is a common technique; we adopt a similar template-based approach as in}~\cite{liu2024routing}.
Each cluster is matched separately, and the influence between different clusters is not considered in this step.

\minisection{Contour-based Template Generation}
\revise{The template grids are generated based on the bounding box of the contour as follows:
\begin{enumerate}
    \item \textbf{Grid Cell Generation:} We tile the bounding box of the contour with a uniform grid. Each grid cell has dimensions equal to the maximum macro width $w_{\max}$ and maximum macro height $h_{\max}$ within the cluster, plus half of the minimum channel spacing $c$ on each side, i.e., cell size is $(w_{\max} + c, h_{\max} + c)$. This ensures that when macros are placed at grid positions, they maintain proper spacing.
    \item \textbf{Distance Calculation:} For each grid cell center, we compute its signed Euclidean distance to the nearest point on the contour boundary (using shapely library). A negative distance indicates the cell center is   inside the contour; a positive distance indicates it is outside.
    \item \textbf{Grid Selection:} We rank all grid cells by their signed distance (from most negative to most positive) and select the top-$k$ cells, where $k$ equals $n$, the number of macros in the cluster. This preferentially selects cells inside or near the contour boundary, allowing some flexibility to accommodate macros slightly outside the contour if needed.
    \item \textbf{Feasibility Filtering:} We discard any grid position that would cause a macro to violate die‑area boundaries or overlap fixed obstacles.
\end{enumerate}
Note that $k$ and $n$ in the assignment problem formulation refer to the same quantity: the number of macros in the cluster to be legalized.}

\minisection{Macro-to-Grid Assignment}
We see legalizing macros to the template grids while minimizing the displacement as an assignment problem on a bipartite graph, 
where nodes on one side represent all macros in the cluster, and nodes on the other side represent available grid positions in the template. 
For each macro-grid position pair, we define the edge weight as the Manhattan distance from the macro's initial position to the grid position. 
We formulate this as a linear assignment problem:
\begin{equation}
    \begin{array}{rll}
        \min&\multicolumn{2}{l}{\displaystyle\sum_{m,g} c_{m,g}\cdot x_{m,g}}\\[4\jot]
        \text{s.t.}&\displaystyle\sum_{j=1}^n x_{ij} = 1, &\forall i \in \{1, 2, \dots, n\}  \\[4\jot]
        & \displaystyle\sum_{i=1}^n x_{ij} = 1, &\forall j \in \{1, 2, \dots, n\}  \\[4\jot]
        & x_{ij} \in \{0, 1\}, &\forall i, j \in \{1, 2, \dots, n\} 
    \end{array}
\end{equation}
where $c_{m,g}$ is the Manhattan distance from the macro's $m$ initial position to the grid position $g$, $x_{m,g}$ is the assignment variable.
By solving this linear assignment problem, we obtain an optimal macro-to-grid assignment that minimizes total displacement. We employ the Hungarian algorithm to solve it.

The whole flow for template-based macro matching is shown in \Cref{figs:template_matching}. Consider the initial macro placement solution generated by mixed-size placement. First, we use the LLM Agent-generated contour algorithm to generate various contours. Subsequently, our designed  template  generation algorithm produces a template grid based on these contours. Then, the linear assignment problem is solved to assign macros to  grids. Finally, multiple macro placement results are generated for each cluster.
Here, if only one solution is needed, we can choose the solution with the smallest displacement or the one with the greatest regularity.
In our experiments, we chose the solution with the smallest displacement, since it is hard to characterize the trade-off between regularity and displacement. However, the results generated by MacroAgent already improve the regularity of the initial mixed-size placement a lot.
Since LLM can efficiently find a lot of heuristics to fit the real macro legalization testcases, we can obtain an exponential number of  solutions for the overall legalization through combination. In the experiment, by sorting the displacements of candidates in each cluster, we quickly identified the global macro legalization candidates (after intra-cluster legalization) with the top K \revise{smallest} displacements. This \textbf{multiple solution capability can increase the robustness of legalization} (find a legal macro legalization).

\subsection{Inter-Cluster Refinement}

Since each cluster is optimized independently by the LLM to maximize local regularity, 
the resulting contours may overlap with neighboring clusters. 
The Inter-Cluster Refinement stage~\Cref{eq:macro_legalization_tcg} is specifically designed to resolve these global conflicts 
while strictly preserving the internal relative positions (regularity) achieved by the LLM in each cluster.

Our approach adapts the traditional constraint graph formulation~\cite{chen2023stronger} to handle clusters with irregular shapes. 
\begin{equation}
\label{eq:macro_legalization_tcg}
\begin{array}{rll}
\min & \multicolumn{2}{l}{\displaystyle\left\|\revise{\vec{x}} - \revise{\vec{x}^{\prime}}\right\|_1 + \left\|\revise{\vec{y}} - \revise{\vec{y}^{\prime}}\right\|_1} \\[\jot]
\text{s.t.} & \displaystyle x_i + w_i \leq x_j, & \forall e_{ij} \in G_x \\[\jot]
& \displaystyle y_i + h_i \leq y_j, & \forall e_{ij} \in G_y \\[\jot]
& \multicolumn{2}{l}{\displaystyle W_l \leq x_i \leq W_h - w_i}\\[\jot]
& \multicolumn{2}{l}{\displaystyle H_l \leq y_i \leq H_h - h_i} \\[\jot]
& \displaystyle x_{a_i^k} - x_{a_j^k} = x_{a_i^k}' - x_{a_j^k}', & \forall a_i^k, a_j^k \in {C}^k
\end{array}
\end{equation}

As shown in \Cref{eq:macro_legalization_tcg}, we formulate a linear programming (LP) problem that minimizes the total displacement $\| \revise{\vec{x}} - \revise{\vec{x}^{\prime}} \|_1 + \| \revise{\vec{y}} - \revise{\vec{y}^{\prime}} \|_1$ between the original positions $(\revise{\vec{x}^{\prime}}, \revise{\vec{y}^{\prime}})$ and the legalized positions  $(\revise{\vec{x}}, \revise{\vec{y}})$. The first and second constraints enforce non-overlapping conditions through horizontal and vertical constraint graphs $G_x$ and $G_y$, while the third and fourth constraints ensure that all macros remain within the placement boundaries ($W_l$ for x coordinate of lower left point, $W_h$ for x coordinate of upper right point, $H_l$ for y coordinate of lower left point, $H_h$ for y coordinate of upper right point). The only difference with~\cite{chen2023stronger} is the last constraint $x_{a_i^k} - x_{a_j^k} = x_{a_i^k}' - x_{a_j^k}'$ preserves the relative positioning of macros within the same cluster ${C^k}$, thus maintaining the regularity.

For scenarios where the linear programming approach fails to solve (due to extra constraints), we employ DREAMPlace's default heuristic algorithms~\cite{lin2020dreamplace} as a fallback mechanism. Although this approach may compromise the achieved regularity, it provides an efficient solution for resolving overlaps in practice.

\section{Implementation Details}
\label{sec:impl}
The macro legalization algorithm was implemented in Python. This was due to LLMs' generally better performance with interpreted languages like Python. 

We leverage cursor, Cua~\cite{cua} and pyautogui for constructing the Agent workflow. Cursor is a general AI agent editor, and Cua and pyautogui are used to simulate the mouse and keyboard to control the cursor editor to automate the flow. We use Cua and pyautogui to program the fixed flow.
We install the \texttt{sequential thinking}, \texttt{memory}, and \texttt{context7} mcp in the cursor, 
endowing the agent with capabilities of thinking, searching and memory. 
After seeding the template prompt, the agent runs fully automatically: Cursor orchestrates file edits and execution; Cua and pyautogui drive the UI; failure handling (signature mismatch, runtime error, invalid shape) is automated via the corresponding validators. No human edits are applied to candidate code during the search.

\revise{We prioritized models that reliably follow structured code-generation instructions and produce long,
compositional functions with few syntax errors---key requirements for agentic code generation (see \Cref{sec:llm_agents}).
In our budgeted setting, Claude 4.0 Sonnet (Thinking, Max) best satisfied these criteria.}
We also spot-checked Gemini 2.5 Pro and GPT 5 and observed comparable adherence to the prompt scaffolding; a smaller 30-40B-class model struggled with syntax/structure. 
A formal cross-model benchmark is valuable but outside our scope; 
we therefore document the budgeted agent setting. \revise{The macro legalization source code is publicly available at \url{https://github.com/gilgamsh/MacroAgent}; the LLM prompts and agent scripts will be released in the same repository upon acceptance to facilitate reproduction and cross-model comparisons.}
\section{Experiment}
\label{sec:experiment}

\subsection{Experiment Setup}
\minisection{Baseline Selection}
DREAMPlace default macro legalization~\cite{lin2020dreamplace} served as the primary baseline method, utilizing heuristic algorithms and constraint graph techniques for macro legalization. 
Additionally, the sequence pair method presented in~\cite{chen2023stronger} is also included; it uses the sequence pair method and simulated annealing to enhance the robustness of macro legalization. 
Neither of these baseline algorithms explicitly incorporates macro regularity and typically relies on only one or two \revise{heuristics}.

We distinguish the scope of macro legalization from macro placement. While most prior macro placement works focus on optimizing rough locations that allow minor overlaps~\cite{lin2018regularity,agnesina2023autodmp,peng2023pplace,chen2023stronger,xuereinforcement,jiang2025regplace,pu2024incremacro}, and rely on a macro legalizer to resolve overlaps, MacroAgent operates as a downstream refinement framework designed to resolve overlaps and improve regularity of existing placements. Therefore, comparing MacroAgent directly with mixed-size global placement-based macro placement tools (which generate the inputs for our system) is methodologically inconsistent. Our work is orthogonal to prior macro placement works and can be integrated with them. 
Instead, we anchor all methods to the same mixed-size global placement prototype to isolate legalization effects. We validate its performance against state-of-the-art legalization and refinement algorithms (e.g., DREAMPlace's legalization module and sequence pair method in~\cite{chen2023stronger}), ensuring a fair comparison within the same physical design stage.
\revise{We also compare with the macro legalization capability of Cadence Innovus, a leading industrial physical design tool.}

\minisection{Experiment flow}
The experiment workflow starts mixed-size placement using DREAMPlace 4.1~\cite{chen2023stronger}. 
Subsequently, the macro legalization  is applied. 
Next, we utilize DREAMPlace 4.1~\cite{chen2023stronger} to implement standard cell placement and legalization. 
Finally, we operate the HeLEM-GR~\cite{zhao2024helem} for global routing and evaluate the performance of the resulting layout.
\revise{We refer to this as the \emph{academic flow}. We additionally evaluate under an \emph{industrial flow}, where standard cell placement and routing are performed entirely within Cadence Innovus (detailed in \Cref{sec:ppa_eval}).}

\begin{figure}[tb]
    \centering
    \includegraphics[width=.78\linewidth]{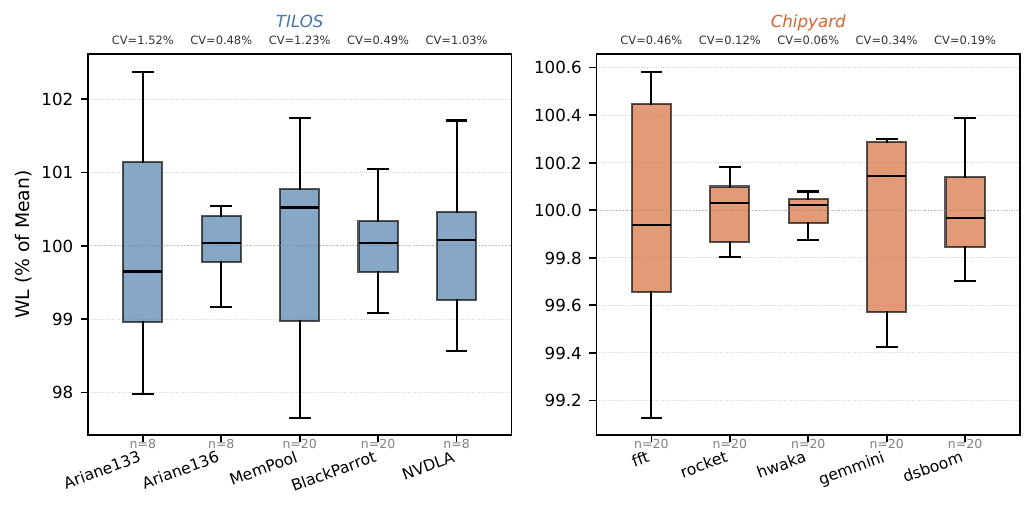}
    \caption{\revise{Normalized wirelength distribution of MacroAgent across all candidate results per testcase.}}
    \label{fig:boxplot_stability}
\end{figure}

The experiment was configured on a Linux machine \revise{running CentOS 7, equipped with an Intel Xeon Platinum 8358 CPU @ 2.60GHz, 1TB of RAM, and an NVIDIA A800-SXM4-80GB GPU (CUDA 12.1). The software environment includes GCC 9.5.0 and Python 3.9.18.}
The proposed macro legalization algorithm exclusively utilizes CPU resources, while GPU resources are leveraged by DREAMPlace and HeLEM-GR.
\revise{Cadence Innovus v22.10 is used for the industrial baseline comparison and PPA evaluations.}
Due to the inherent randomness in sequence pair legalization, we execute the legalization algorithm 20 times with different random seeds. To ensure fairness, MacroAgent generates at most 20 legalization candidate results that minimize displacements. (note that MacroAgent can generate diverse legalization results as needed, \textbf{multiple solution capability}) For each result, we perform standard cell placement and global routing. Finally, the result table lists the best wirelength-based outcomes, along with their corresponding legalization runtime, total runtime, and other relevant metrics (total 20 times runtime for sequence pair and MacroAgent).
\revise{To demonstrate the stability of MacroAgent, \Cref{fig:boxplot_stability} shows the distribution of routed wirelength across all candidate results for each testcase, normalized by the per-testcase mean.
MacroAgent enumerates candidate solutions in order of increasing displacement, so the resulting variations primarily stem from permutations within small macro clusters, while the placement of large clusters remains stable.
Consequently, testcases with diverse macro types (e.g., Chipyard, 7--12 types) exhibit particularly tight distributions (CV $< 0.5\%$), as the dominant clusters are effectively fixed and only minor clusters are permuted.
TILOS testcases with a single macro type show slightly higher yet still modest variation (CV $\approx 1\%$--$1.5\%$), because the single-cluster structure exposes more of the layout to reordering.
Overall, the best-of-$N$ selection introduces negligible cherry-picking bias.}

\minisection{Benchmark}
For benchmarking, the TILOS~\cite{cheng2023assessment} benchmark was chosen\revise{, following the same testcase configuration as~\cite{chen2023stronger}: Ariane133, Ariane136, and MemPool use the ASAP7 PDK, while BlackParrot and NVDLA use the NanGate45 PDK}.
Furthermore, we found that TILOS is an easy benchmark for macro legalization and that all baseline methods can legalize all testcases.
To model more challenging macro legalization scenarios, several macro circuit designs generated via Chipyard~\cite{chipyard} \revise{(v1.9.1) with the ASAP7 PDK} were also incorporated into the benchmarks.
The statistics of the TILOS and Chipyard testcases are shown in \Cref{tab:tilos_stats} and \Cref{tab:chipyard_stats}.
$\#\text{macro}$ is the number of macros, $\#\text{types}$, $\#\text{inst}$, and  $\#\text{net}$ denote the number of macro types, the number of instances,  and the number of nets.
\textbf{We can see that the Chipyard testcases are more challenging, having more macros and a wider range of macro types.}

\begin{table}[tb!]
    \centering
    \footnotesize
    \caption{Benchmark statistics of TILOS testcases (easy).}
    \label{tab:tilos_stats}
    \begin{tabular}{|c|cccc|}
        \hline
        Testcase & \#macro & \#types & \#inst & \#net \\ \hline \hline
        \texttt{Ariane133} & 133 & 1 & 98250 & 101335 \\
        \texttt{Ariane136} & 136 & 1 & 142840 & 144683 \\
        \texttt{MemPool} & 20 & 2 & 131732 & 135299 \\
        \texttt{BlackParrot} & 220 & 6 & 1277012 & 1510338 \\
        \texttt{NVDLA} & 128 & 1 & 153561 & 208063 \\ \hline
    \end{tabular}
\end{table}

\begin{table}[tb!]
    \centering
    \footnotesize
    \caption{Benchmark statistics of Chipyard testcases (hard).}
    \label{tab:chipyard_stats}
    \begin{tabular}{|c|cccc|}
        \hline
        Testcase & \#macro & \#types & \#inst & \#net \\ \hline \hline 
        \texttt{fft} & 121 &7 & 211905 & 216507 \\
        \texttt{hwaka} & 292 & 9& 680029 & 687204 \\
        \texttt{gemmini} & 737 &9 & 1176503 & 1189717 \\
        \texttt{dsboom} & 556 & 12 & 643392 & 659154 \\
        \texttt{rocket} & 121 & 7& 203995 & 208595 \\ \hline
    \end{tabular}
\end{table}


\begin{table*}[tb]
    \centering
    \setlength{\tabcolsep}{1.2pt} 
    \renewcommand{\arraystretch}{1.4}
    \caption{Results on TILOS \revise{(academic flow: DREAMPlace + HeLEM-GR)}: regularity, routed wirelength\revise{,} congestion (\%) and runtime (s, sum of 20 runs). S (Status) shows if failed to legalize.}
    \label{tab:tilos_results}
    \resizebox{\linewidth}{!}{
    \begin{tabular}{|c|cccccc|cccccc|cccccc|}
        \hline
        \multirow{2}{*}{Testcase} & \multicolumn{6}{c|}{DREAMPlace~\cite{lin2020dreamplace}} & \multicolumn{6}{c|}{Sequence Pair~\cite{chen2023stronger}} & \multicolumn{6}{c|}{MacroAgent} \\
        & S & Reg & LT & WL & Con & TT & S & Reg & LT & WL & Con & TT & S & Reg & LT & WL & Con & TT \\
        \hline \hline
        \texttt{Ariane133} & $\checkmark$ & 0.13 & 0.004 & 8.69E08 & 0.03 & 59.2 & $\checkmark$ & 0.16 & 38.625 & 8.51E08 & 0.23 & 376.7 & $\checkmark$ & 0.67 & 1.570 & \textbf{7.62E08} & \textbf{0.07} & 130.9 \\
        \texttt{Ariane136} & $\checkmark$ & 0.13 & 0.005 & 8.37E08 & 0.02 & 90.1 & $\checkmark$ & 0.18 & 93.652 & 8.44E08 & 0.00 & 576.9 & $\checkmark$ & 0.59 & 1.560 & \textbf{8.35E08} & \textbf{0.00} & 188.4 \\
        \texttt{MemPool} & $\checkmark$ & 0.13 & 0.001 & 7.46E08 & 0.77 & 85.0 & $\checkmark$ & 0.13 & 1.077 & 7.44E08 & 0.48 & 557.4 & $\checkmark$ & 0.35 & 1.050 & \textbf{6.80E08} & \textbf{0.39} & 501.6 \\
        \texttt{BlackParrot} & $\checkmark$ & 0.06 & 0.008 & 9.10E10 & 3.90 & 503.7 & $\checkmark$ & 0.06 & 254.212 & \textbf{9.05E10} & \textbf{3.41} & 3930.7 & $\checkmark$ & 0.13 & 5.550 & 9.19E10 & 4.28 & 3850.2 \\
        \texttt{NVDLA} & $\checkmark$ & 0.08 & 0.004 & 2.46E10 & 4.58 & 105.5 & $\checkmark$ & 0.08 & 87.272 & 2.42E10 & \textbf{4.12} & 791.2 & $\checkmark$ & 0.64 & 3.470 & \textbf{2.36E10} & 4.23 & 329.2 \\
        \hline \hline
        Ratio &  & 1.00 & 1.00 & 1.00 & 1.00 & 1.00 & &  1.12 & 16611.63 & 0.99 & 2.01 & 6.93 &  & 4.51 & 663.15 & \textbf{0.95} & \textbf{0.97} & 4.19 \\
        \hline
    \end{tabular}
    }
\end{table*}


\begin{table*}[tb]
    \centering\small
    \setlength{\tabcolsep}{1.2pt} 
    \renewcommand{\arraystretch}{1.4}
    \caption{Results on Chipyard \revise{(academic flow: DREAMPlace + HeLEM-GR)}: regularity, routed wirelength\revise{,} congestion (\%) and runtime (s, sum of 20 runs).  S (Status) shows if failed to legalize.}
    \label{tab:chipyard_results}
    \resizebox{\linewidth}{!}{
    \begin{tabular}{|c|cccccc|cccccc|cccccc|}
        \hline
        \multirow{2}{*}{Testcase} & \multicolumn{6}{c|}{DREAMPlace~\cite{lin2020dreamplace}} & \multicolumn{6}{c|}{Sequence Pair~\cite{chen2023stronger}} & \multicolumn{6}{c|}{MacroAgent} \\
        & S & Reg & LT & WL & Con & TT & S & Reg & LT & WL & Con & TT & S & Reg & LT & WL & Con & TT \\
        \hline \hline
        \texttt{fft} & \cellcolor{red!25}$\times$ & 0.09 & 0.004 & 3.16E10 & 0.33 & 111.3 & \cellcolor{red!25}$\times$ & 0.09 & 2.576 & 3.16E10 & 0.34 & 623.7 & $\checkmark$ & 0.34 & 2.740 & \textbf{3.50E10} & \textbf{0.56} & 656.7 \\
        \texttt{hwaka} & \cellcolor{red!25}$\times$ & 0.06 & 0.015 & 8.40E10 & 0.29 & 357.5 & $\checkmark$ & 0.06 & 535.677 & \textbf{8.38E10} & 0.32 & 2442.0 & $\checkmark$ & 0.14 & 6.440 & 8.50E10 & \textbf{0.29} & 2199.8 \\
        \texttt{gemmini} & $\checkmark$ & 0.05 & 0.029 & 1.10E11 & 0.14 & 354.5 & $\checkmark$ & 0.05 & 413.445 & 1.10E11 & 0.14 & 2202.9 & $\checkmark$ & 0.20 & 14.630 & \textbf{1.07E11} & \textbf{0.13} & 2236.5 \\
        \texttt{dsboom} & $\checkmark$ & 0.05 & 0.019 & 7.89E10 & \textbf{0.11} & 533.8 & $\checkmark$ & 0.05 & 409.577 & 7.89E10 & 0.12 & 2404.7 & $\checkmark$ & 0.15 & 9.110 & \textbf{7.62E10} & 0.18 & 2472.8 \\
        \texttt{rocket} & \cellcolor{red!25}$\times$ & 0.10 & 0.006 & 3.07E10 & 0.16 & 862.1 & $\checkmark$ & 0.10 & 1.922 & 3.36E10 & 0.16 & 862.1 & $\checkmark$ & 0.26 & 2.820 & \textbf{2.87E10} & \textbf{0.11} & 928.3 \\
        \hline
    \end{tabular}
    }
\end{table*}

\subsection{\revise{Prompt Design}}
\revise{To enhance reproducibility, \Cref{figs:prompt_e2} presents a complete E2 (exploration with synthesis) prompt example used in the contour algorithm discovery. The prompt is organized into the six sections described in \Cref{sec:algo}: task description, context with previous results, evolution directive, usage scenario, quality metrics, and input references. Slot variables (shown in braces) are automatically populated from the algorithm database at each generation. We choose E2 as the representative example because it is the most comprehensive strategy, requiring the LLM to first summarize common ideas across five parent algorithms and then synthesize a substantially different new heuristic. The \colorbox{yellow!20}{highlighted} evolution directive is the only section that differs across the five strategies (E1/E2/M1--M3); all other sections remain identical across strategies.}

\begin{tcolorbox}[
  colback=white, colframe=black,
  title={\textbf{E2 Prompt Example for Contour Algorithm Generation}},
  fonttitle=\small\bfseries,
  boxrule=0.8pt, arc=2pt,
  left=4pt, right=4pt, top=2pt, bottom=2pt,
  breakable
]
\small
\textbf{\#\# Task} \\
Design and implement a contour generation algorithm that, given a cluster of macros, produces a closed polygon enclosing them. The algorithm must be implemented as a Python function with the following signature: \\
\texttt{def generate\_contour(corners: list[tuple], centers: list[tuple], polygons: List[Polygon]) -> Polygon} \\
where \texttt{corners} is a list of (x,y) coordinates of all macro corners in the cluster, \texttt{centers} is a list of (x,y) coordinates of macro centers, \texttt{polygons} is a list of Shapely \texttt{Polygon} objects representing each macro's bounding box, and the return value is a Shapely \texttt{Polygon}. \\[4pt]

\textbf{\#\# Context} \\
\textbf{Previous method ideas}: \texttt{\{ideas\}} \\
\textbf{Previous results}: \texttt{\{displacement\}}, \texttt{\{regularity\_index\}}, \texttt{\{solutions\}} and \texttt{\{visualization\}} \\[4pt]

\colorbox{yellow!20}{\textbf{Evolution Directive [E2]:}} \\
I have 5 existing contour generation algorithms with their ideas and codes as follows: \\
\textit{No.1 Idea: Compute alpha shape using Delaunay triangulation with adaptive circumradius filtering...} \\
\textit{Code:} \texttt{def generate\_contour(corners, centers, ...): ...} \\
... \\
\textit{No.5 Idea: Grid-based rectilinear boundary tracing with density-aware cell merging...} \\
\textit{Code:} \texttt{def generate\_contour(corners, centers, ...): ...} \\[2pt]
Please help me create a new contour algorithm that is different from the given ones but can be motivated by them. \\
\textbf{Firstly,} identify the common idea in the provided algorithms. \\
\textbf{Secondly,} based on the backbone idea, describe your new algorithm in one sentence. \\
\textbf{Thirdly,} implement it as a Python function named \texttt{generate\_contour} following the signature defined above. \\[4pt]

\textbf{\#\# Usage Scenario} \\
In the MacroAgent framework, the generated contour serves as the boundary for template generation: the contour polygon is converted into a rectilinear template that defines legal macro slots, and the Hungarian algorithm then assigns macros to these slots to minimize displacement. A contour that closely follows the original macro arrangement preserves macro positions and reduces displacement, while a contour with fewer concavities produces a more regular template with fewer channels. These two objectives often conflict, and the contour algorithm must balance them. The contour algorithm is therefore the key heuristic component whose quality directly determines the downstream legalization result. \\[4pt]

\textbf{\#\# Quality Metrics} \\
Regularity: measured by the channel count ratio $R = C_{\min} / C_{\text{actual}}$, where $C_{\min}$ is the theoretical minimum number of channels for a perfect grid arrangement and $C_{\text{actual}}$ is the actual channel count computed from the Hanan grid. $R \in (0, 1]$; higher is better ($R=1$ means a perfect regular array). \\
Displacement: total Manhattan distance $\sum_i (|x_i - x_i'| + |y_i - y_i'|)$ between each macro's legalized position $(x_i, y_i)$ and its original placement position $(x_i', y_i')$; lower is better. \\[4pt]

\textbf{\#\# Input and Reference} \\
\texttt{\{test case files and reference legalization visualizations\}}
\end{tcolorbox}
\captionof{figure}{\revise{Complete E2 (exploration with synthesis) prompt example for contour algorithm generation. The \colorbox{yellow!20}{highlighted} evolution directive section is the only part that differs across the five strategies (E1/E2/M1--M3); all other sections remain identical. Slot variables (in braces) are populated from the algorithm database at each generation.}}
\label{figs:prompt_e2}

\subsection{Agent Discovered Contour Algorithms}
We use the easy TILOS testcases to generate the contour algorithms, and the hard Chipyard testcases to test the \textbf{generalization ability} of the contour algorithm we discovered.
Meanwhile, we extract the single-cluster testcases decomposed from TILOS testcases to simplify the problem instance input to the LLM.
The LLM-driven algorithm optimization finishes in one day, showing significant efficiency gains versus researchers' development time.
\revise{The discovery run is an offline, one-time search.}
\revise{Among the 110 generated candidates, only 8 algorithms remain in the final database because they improve either displacement or regularity on at least one testcase; the rest are discarded during screening.}
\revise{Notably, nearly all candidates are executable and produce valid results. This confirms that our domain-agnostic abstraction effectively reduces the problem to pure geometry reasoning, which is well within the LLM's capability.}

\revise{In the initialization stage, the LLM generates $N=10$ diverse contour heuristics from scratch based solely on the input prompt, without any hand-crafted algorithm design. These heuristics span four broad geometric strategy families: triangulation-based (Items 1, 2, 5), graph-based (Items 4, 10), density/field-based (Items 6, 7, 8), and grid/scanline-based (Items 3, 9):
\begin{enumerate}
    \item \textbf{Alpha Shape} — adaptive Delaunay triangulation with circumradius filtering.
    \item \textbf{kNN Concave Hull} — locally adaptive concave boundary via kNN density estimation.
    \item \textbf{Grid-Based} — occupancy grid dilation with convex hull extraction.
    \item \textbf{MST} — buffered minimum spanning tree of macro centers.
    \item \textbf{Corner-Preserving Alpha} — alpha shape augmented with macro corner points.
    \item \textbf{KDE Contour} — Gaussian kernel density iso-contour extraction.
    \item \textbf{Voronoi Clip} — Voronoi cells clipped by local neighborhood radius.
    \item \textbf{Relative Position} — kNN-biased alpha shape preserving local structure.
    \item \textbf{Rectilinear Boundary Tracing} — scanline-based orthogonal interval merging.
    \item \textbf{Manhattan Distance Hull} — $L_1$-weighted MST with axis-aligned buffering.
\end{enumerate}
Detailed implementation is available at \url{https://github.com/gilgamsh/MacroAgent}.
This diversity of geometric primitives ensures that the initial population covers a wide region of the algorithm design space, providing a broad foundation for the subsequent evolutionary search.}

At test time, each cluster generates 8 candidate legalization results; we can directly find the potential minimum displacement legalization result
 by combining all clusters' minimum displacement results. 
 By exploiting the multiple-solution capability, we can find the global macro legalization candidates with the top K smallest displacements. 
 Note that here the displacement is just an estimation, and we still need to perform inter-cluster refinement to get the final legalization result.
 
As shown in \Cref{tab:tilos_results} and \Cref{tab:chipyard_results}, MacroAgent outperforms the baselines on both benchmarks (especially on test benchmark Chipyard),
showing the generalization ability of the contour algorithms we discovered.
\begin{figure}[tb!]
    \centering
    \includegraphics[width=\linewidth]{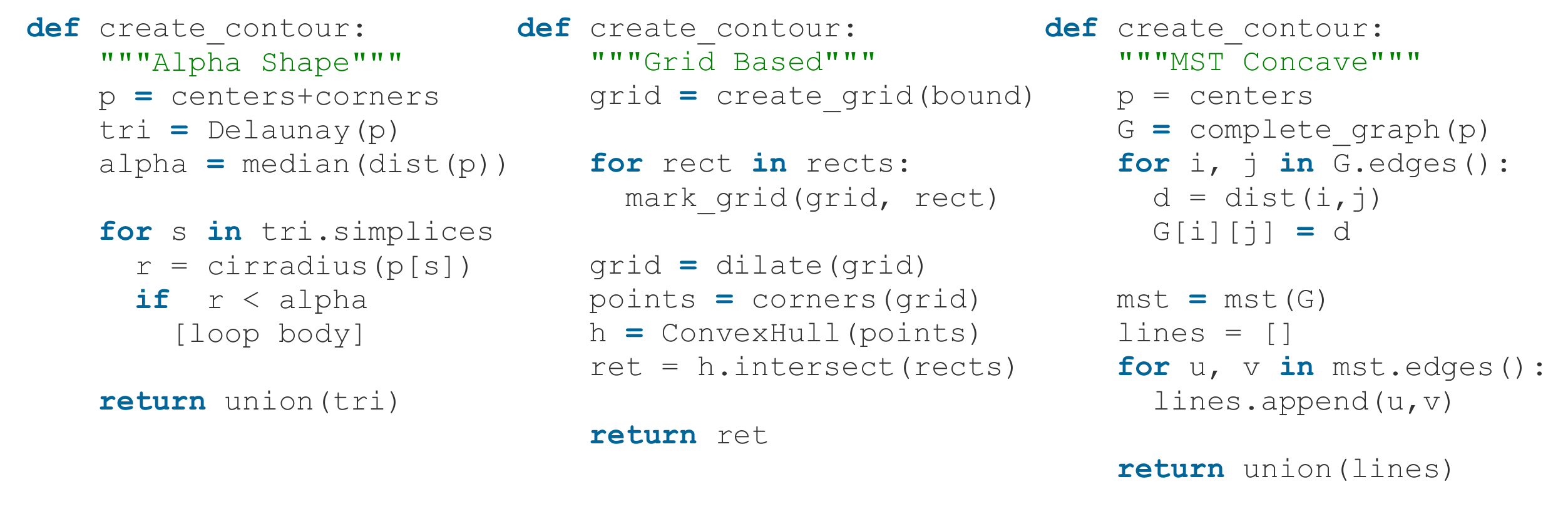}
    \caption{\revise{Selected LLM-designed Contour Algorithms (simplified). Simplified pseudocode of three representative contour generation algorithms discovered by the LLM agent: Alpha Shape (concave boundary via Delaunay triangulation), Grid-Based (occupancy grid with dilation and convex hull), and MST-Based (minimum spanning tree with buffer expansion).}}
    \label{figs:algo}
\end{figure}
\textbf{
Here, we analyze some of the contour algorithms LLM discovered as shown in \Cref{figs:algo}.
}
\revise{
\begin{description}
    \item[\textbf{Alpha Shape}] generates the contour as follows: (1) collect corner and center points of all macros; (2) perform Delaunay triangulation on these points; (3) for each triangle, compute its circumradius; (4) discard triangles whose circumradius exceeds an adaptive threshold---a large circumradius indicates an elongated, non-compact triangle that likely spans empty space between macros; (5) merge the remaining compact triangles to form the contour. This yields a tighter, concave boundary as shown in the third figure in \Cref{fig:viz_combined}.
    \item[\textbf{Grid Based}] generates the contour as follows: (1) compute the bounding box of all macros and partition it into an $N\times N$ uniform grid; (2) for each grid cell, test whether its center point lies inside any macro polygon and mark it as occupied if so; (3) dilate the occupied cells by marking the eight immediate neighbors of every occupied cell; (4) extract the four corner vertices of every marked cell and compute the convex hull; (5) apply light post-processing to obtain the final contour. Since it is based on a grid structure, it produces more regular contours than the other algorithms, as shown in the second figure in \Cref{fig:viz_combined}.
    \item[\textbf{MST} (minimum-spanning tree)] generates the contour as follows: (1) construct a complete graph on macro centers with \textup{Euclidean} edge weights; (2) compute the minimum spanning tree (MST); (3) convert MST edges into line segments and apply a buffer to the segment set; (4) union the buffered region with all macro polygons to obtain the final contour. This method naturally fits the outline of the original macros, with only a simple effect of smoothing the boundaries, as shown in the last figure in \Cref{fig:viz_combined}.
\end{description}
}
From the above analysis, we can see that the LLM designed contour algorithms can achieve different trade-offs between regularity and displacement in algorithm level, not only in parameter level. \textbf{These algorithms are meaningful and readable.}
We also find that with the help of reference manual legalization solutions, LLMs can generate more focused thoughts and modifications.

\begin{figure}[tb]
    \centering
    \includegraphics[width=0.60\linewidth]{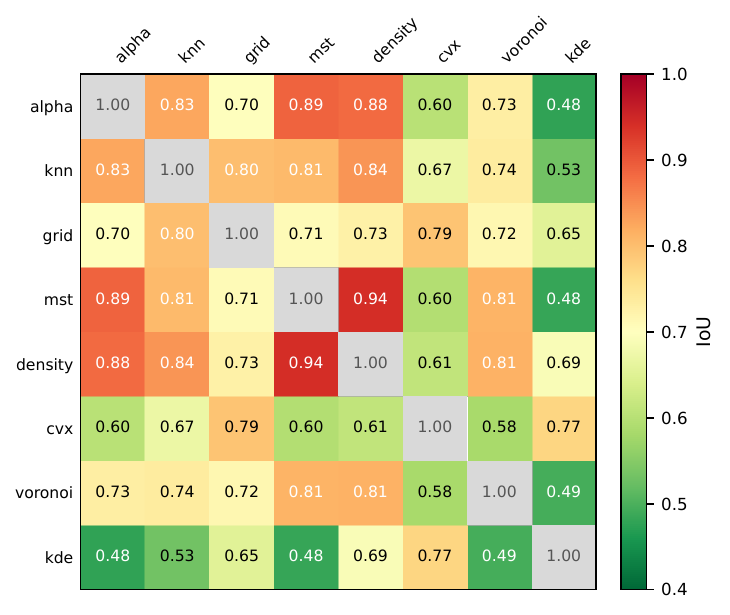}
    \caption{\revise{Pairwise IoU matrix of eight contour algorithms on \texttt{Ariane133}. Lower off-diagonal values indicate higher geometric diversity.}}
    \label{fig:contour_iou}
\end{figure}

\revise{
\minisection{Quantitative Diversity Analysis}
To quantitatively evaluate the diversity of the eight LLM-discovered contour algorithms, we compute the pairwise Intersection-over-Union (IoU) of the contour polygons generated by each algorithm on a representative testcase (\texttt{Ariane133}).
A lower IoU between two algorithms indicates that they produce geometrically distinct contours, confirming genuine algorithmic diversity rather than minor parametric variations.
\Cref{fig:contour_iou} reports the full $8\times 8$ pairwise IoU matrix.
The average off-diagonal IoU is $0.71$. Because all contours enclose the same set of macros, a baseline overlap is inherent; the meaningful variation lies in boundary strategies, where the algorithms differ significantly.
Notably, \texttt{kde} exhibits the lowest average pairwise IoU with all other methods ($0.48$--$0.77$), confirming that it explores a fundamentally different geometric strategy.
While some method pairs share higher similarity (e.g., \texttt{mst} and \texttt{density\_weighted} at $0.94$; \texttt{alpha\_shape} and \texttt{mst} at $0.89$), the eight algorithms span a wide range of contour shapes, which directly enables the multiple-solution capability described above.
}

\subsection{Overall Comparisons}
Regularity reduces channel count and deadspace, which in turn reduces detours and router effort. We therefore report routed wirelength and congestion as PPA-relevant surrogates in our legalization only study.
We list the results of the TILOS and Chipyard testcases in \Cref{tab:tilos_results} and \Cref{tab:chipyard_results}. 
$S$ shows whether the macro legalization is successful. Reg shows the regularity score of the macro legalization. LT shows the time of macro legalization. WL shows the routed wirelength \revise{reported by HeLEM-GR~\cite{zhao2024helem} in DREAMPlace internal units}. Con shows the congestion(\%). Regularity (Reg) is calculated using the channel-based metric defined in~\Cref{equ:regularity_index}, where a score of 1.0 represents a theoretically perfect square array. TT shows the total workflow time(s) (sum of 20 runs), including mixed-size placement, macro legalization, standard cell placement and  legalization, and global routing. 

For the TILOS benchmark, MacroAgent achieves an average 4\% improvement in wirelength and reduces congestion in some testcases compared with sequence pair. Compared with DREAMPlace default macro legalization, MacroAgent delivers a 5\% improvement in wirelength and similar congestion. 
Beyond standard metrics, the robustness of MacroAgent significantly surpasses other methods, particularly in complex design scenarios. As evidenced in~\Cref{tab:chipyard_results}, the DREAMPlace baseline fails to legalize three out of five Chipyard testcases (`fft', `hwaka', `rocket'), and the Sequence Pair method fails on `fft' despite 20 restart attempts. These failures typically stem from the limitation of utilizing a single heuristic, which often leads to getting trapped in local optima or failing to resolve dense overlaps in highly constrained regions. In contrast, MacroAgent leverages a portfolio of diverse LLM-designed heuristics. By generating multiple candidate contours and templates for each cluster, our framework effectively avoids the rigidity of single-algorithm approaches. This capability ensures design closure even in high-density cases.

Furthermore, as shown in \Cref{tab:pair_wise}, MacroAgent achieves a wirelength improvement of 3\% to 5\% with similar congestion compared to both DREAMPlace and sequence pair on the common success testcases. \revise{Note that for failed cases (marked with $\times$ in \Cref{tab:chipyard_results}), the reported wirelength and congestion values are obtained by proceeding with the downstream flow despite remaining macro overlaps; these results are not physically valid and are included only for reference---they should not be used for quantitative comparison.}

The legalization efficiency of MacroAgent is higher than sequence pair (since sequence pair use simulated annealing), with a runtime on the order of seconds and its time consumption in the full workflow being negligible.   \textbf{Overall our MacroAgent framework is more robust, effective with better solution quality.}

\begin{table}[tb]
    \centering\small
    \caption{Metric ratio comparison of MacroAgent and baselines on Chipyard.}
    \label{tab:pair_wise}
    \begin{tabular}{|c|ccccc|}
        \hline
        Comparison & Reg & LT & WL & Con & TT  \\
        \hline\hline
        MacroAgent / DREAMPlace    & 3.50 & 491.98   & 0.97 & 1.28 & 5.47  \\
        MacroAgent / Sequence Pair & 2.98 & 0.38     & 0.95 & 1.01 & 1.01  \\
        \hline
    \end{tabular}
\end{table}

\subsection{\revise{PPA Evaluation}}
\label{sec:ppa_eval}
\revise{
To validate the downstream impact of macro legalization on physical design quality,
we perform complete place-and-route using Cadence Innovus on the TILOS benchmark, where all baseline methods successfully legalize every testcase, enabling a fair PPA comparison. On Chipyard, baseline failures preclude meaningful PPA comparison.
}

\minisection{\revise{Experiment Flow}}
\revise{
After macro legalization, we import the post-legalization DEF and LEF files into Innovus and run \texttt{place\_opt\_design} for standard cell placement and optimization, followed by routing.
We report routed wirelength, WNS (worst negative slack), TNS (total negative slack), and total power.
For each testcase, we set a target frequency (i.e., the reciprocal of the clock period) so that WNS falls within 10--20\% of the clock period.
A negative WNS means the critical path violates the timing constraint; a more negative value indicates a more severe violation.
A positive WNS indicates that all paths meet timing, with a larger value reflecting more slack margin.
In both cases, a higher WNS is preferable.
TNS aggregates all negative slacks across timing endpoints (0 if no negative slack), reflecting the overall timing condition of the design.
}

\begin{table*}[tb]
    \centering
    \setlength{\tabcolsep}{1.2pt}
    \renewcommand{\arraystretch}{1.4}
    \caption{\revise{Comparison of post-route PPA results on TILOS benchmark (industrial flow: Innovus place-and-route): frequency (MHz), routed wirelength (\textmu{m}), WNS (ns), TNS (ns), and power (mW).}}
    \label{tab:tilos_ppa}
    \resizebox{\linewidth}{!}{\color{blue}
    \begin{tabular}{|c|c|cccc|cccc|cccc|}
        \hline
        \multirow{3}{*}{Testcase} & \multirow{3}{*}{Freq} & \multicolumn{4}{c|}{DREAMPlace~\cite{lin2020dreamplace}} & \multicolumn{4}{c|}{Sequence Pair~\cite{chen2023stronger}} & \multicolumn{4}{c|}{MacroAgent} \\
        & & WL & WNS  & TNS  & Power & WL & WNS  & TNS  & Power & WL & WNS  & TNS  & Power  \\
        & (MHz) & (\textmu{m}) & (ns) & (ns) & (mW)  & (\textmu{m}) & (ns) & (ns) & (mW)  & (\textmu{m}) & (ns) & (ns) & (mW)   \\
        \hline \hline
        \texttt{Ariane133} & 500 & 910082 & -0.128 & -2.876 & 235.638 & 907235 & -0.115 & -2.533 & 235.551 & \textbf{883726} & \textbf{-0.066} & \textbf{-0.882} & \textbf{234.954} \\
        \texttt{Ariane136} & 769 & 906339 & -0.021 & -1.554 & 367.128 & 910466 & -0.028 & -1.822 & 367.467 & \textbf{857138} & \textbf{-0.007} & \textbf{-0.060} & \textbf{366.547} \\
        \texttt{MemPool} & 333 & 778501 & -0.396 & -1207.310 & 55.674 & 775212 & -0.395 & -1237.757 & \textbf{55.599} & \textbf{752503} & \textbf{-0.369} & \textbf{-1048.01} & 55.613 \\
        \texttt{BlackParrot} & 667 & 30538061 & 0.144 & 0 & 2072.170 & 30395222 & \textbf{0.150} & 0 & \textbf{2071.562} & \textbf{29831309} & 0.146 & \textbf{0} & 2071.842 \\
        \texttt{NVDLA} & 500 & 8850019 & -0.208 & -2.355 & 1093.415 & 8852443 & -0.222 & -3.275 & 1093.389 & \textbf{8806758} & \textbf{-0.021} & \textbf{-0.129} & \textbf{1092.430} \\
        \hline \hline
        Ratio & - & 1.000 & 1.000 & 1.000 & 1.000 & 0.998 & 1.050 & 1.117 & 0.999 & \textbf{0.971} & \textbf{0.574} & \textbf{0.317} & \textbf{0.998} \\
        \hline
    \end{tabular}
    }
\end{table*}




\minisection{\revise{Results Analysis}}
\revise{
As shown in \Cref{tab:tilos_ppa}, MacroAgent provides the strongest overall PPA trade-off among the three methods.
Compared with DREAMPlace, MacroAgent reduces routed wirelength by 2.9\% on average, with the largest improvement on \texttt{Ariane136} ($-5.4\%$).
The improved regularity reduces deadspace between macros, which shortens routing detours and lowers wirelength; the wirelength reduction in turn improves downstream PPA.
This wirelength reduction directly translates into timing improvements: MacroAgent achieves 42.6\% WNS improvement and 68.3\% TNS improvement on average compared with DREAMPlace.
The gains are particularly notable on \texttt{Ariane136} and \texttt{NVDLA}, where TNS is reduced by over 90\%.
Power consumption remains comparable across all methods, as macro legalization primarily affects interconnect topology rather than cell-level switching activity. MacroAgent achieves lower routed wirelength on all five testcases and better overall timing than Sequence Pair (best WNS on four of five testcases and equal-best TNS on all five), with comparable power.
}

\minisection{\revise{Comparison with Innovus Macro Placement}}
\revise{
To further evaluate MacroAgent against industrial tools, we construct two controlled flows on the five TILOS testcases.
\textbf{Flow A (Innovus baseline):} DREAMPlace initial placement $\rightarrow$ Innovus macro refinement via \texttt{place\_design~-concurrent\_macros~-incremental} followed by \texttt{refine\_macro\_place} (which includes flip optimization) $\rightarrow$ \texttt{place\_design} for standard cell placement $\rightarrow$ routing.
\textbf{Flow B (MacroAgent + Innovus):} DREAMPlace initial placement $\rightarrow$ MacroAgent macro legalization $\rightarrow$ the same \texttt{refine\_macro\_place} and downstream stages as Flow~A.
Both flows share identical Innovus settings and the same downstream pipeline after macro refinement.
Because flip optimization cannot be disabled in \texttt{refine\_macro\_place} and significantly affects wirelength, both flows include it.
Therefore, the only controlled variable is the legalized macro placement entering the shared Innovus pipeline, isolating the impact of macro legalization quality on final PPA.
}

\begin{table}[tb]
    \centering\footnotesize
    \setlength{\tabcolsep}{2.6pt}
    \renewcommand{\arraystretch}{1.15}
    \caption{\revise{Comparison with Innovus macro placement on TILOS (industrial flow: Innovus place-and-route): routed wirelength (\textmu{m}), congestion (\%), and runtime (s).}}
    \label{tab:innovus_comparison}
    {\color{blue}
    \begin{tabular}{|c|ccc|ccc|}
        \hline
        \multirow{2}{*}{Testcase} & \multicolumn{3}{c|}{Innovus} & \multicolumn{3}{c|}{MacroAgent} \\
        & WL (\textmu{m}) & Con (\%) & RT (s) & WL (\textmu{m}) & Con (\%) & RT (s) \\
        \hline \hline
        \texttt{Ariane133} & 727027 & 0.02 & 408 & \textbf{701629} & \textbf{0.01} & \textbf{379} \\
        \texttt{Ariane136} & 893684 & 0.01 & 491 & \textbf{892388} & \textbf{0.01} & \textbf{437} \\
        \texttt{MemPool} & 821082 & 0.01 & 432 & \textbf{810131} & \textbf{0.01} & \textbf{417} \\
        \texttt{BlackParrot} & 28421983 & \textbf{0.16} & 2938 & \textbf{28341650} & 0.29 & \textbf{2635} \\
        \texttt{NVDLA} & 8325486 & 0.63 & 551 & \textbf{8033640} & \textbf{0.61} & \textbf{512} \\
        \hline \hline
        Ratio & 1.000 & 1.00 & 1.00 & \textbf{0.982} & 1.06 & \textbf{0.92} \\
        \hline
    \end{tabular}
    }
\end{table}

\revise{
As shown in \Cref{tab:innovus_comparison}, MacroAgent achieves lower routed wirelength than Innovus on all five testcases, with an average improvement of 1.8\%.
The improvement is most pronounced on \texttt{Ariane133} and \texttt{NVDLA} (both $-3.5\%$).
Since Innovus's macro placement refinement includes flip optimization, it may further adjust macro positions based on orientation changes---an optimization that can benefit certain testcases.
This explains the congestion advantage of Innovus on \texttt{BlackParrot} ($0.16\%$ vs.\ $0.29\%$): Innovus's integrated orientation-flip refinement effectively reduces routing hotspots.
However, flip optimization does not always help: compared to the results without Innovus refinement in \Cref{tab:tilos_ppa}, the wirelength of \texttt{Ariane136} and \texttt{MemPool} increases in both flows, suggesting that the orientation changes can degrade placement quality on certain designs.
Despite this, MacroAgent still achieves consistently lower wirelength across all testcases, indicating that its legalization provides a superior starting point that Innovus's downstream optimization preserves and benefits from.
}

\subsection{Analysis and Visualization}

\begin{figure}[tb]
  \centering\small
  \begingroup
    \setlength{\tabcolsep}{1.5pt}
    \begin{tabular}{cccc}
    \subfloat[Initial]{%
      \centering\includegraphics[width=.25\linewidth]{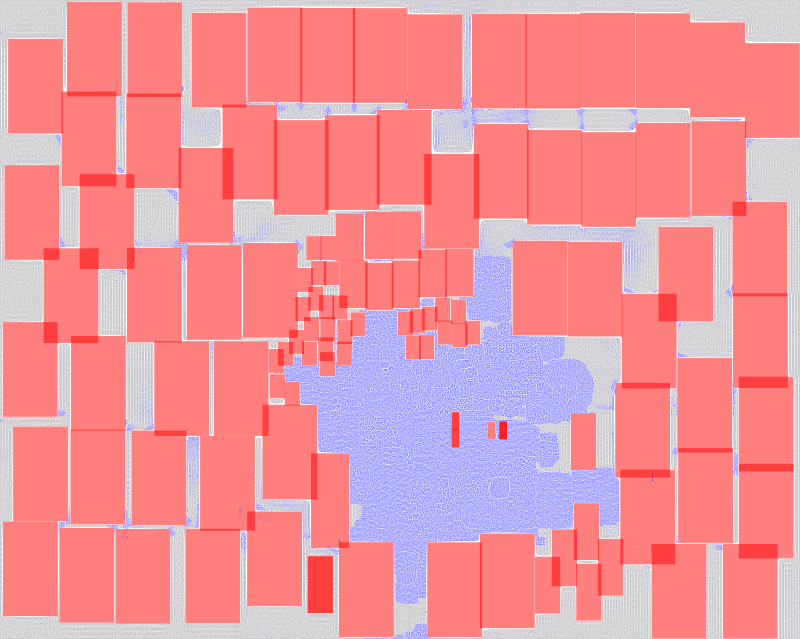}}&
    \subfloat[DREAMPlace]{
      \centering\includegraphics[width=.25\linewidth]{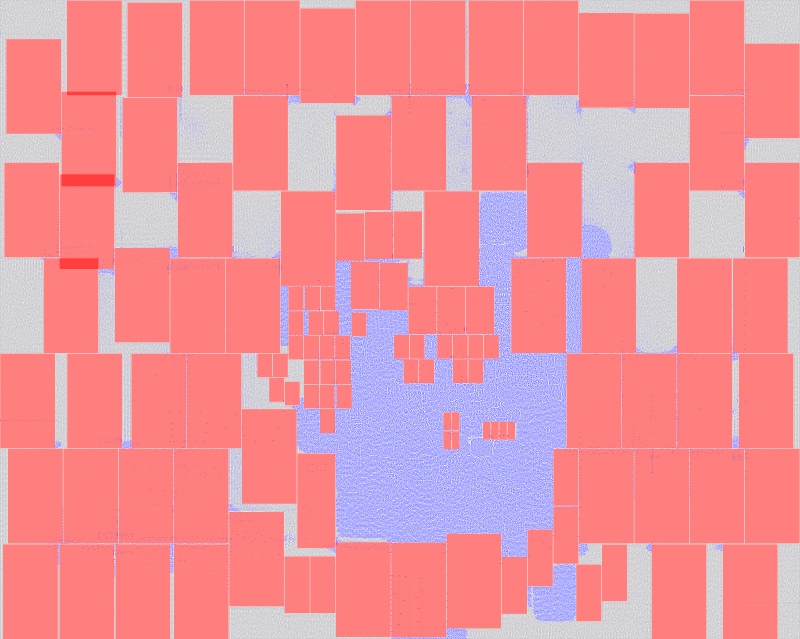}
    }&
    \subfloat[Sequence Pair]{%
      \centering\includegraphics[width=.25\linewidth]{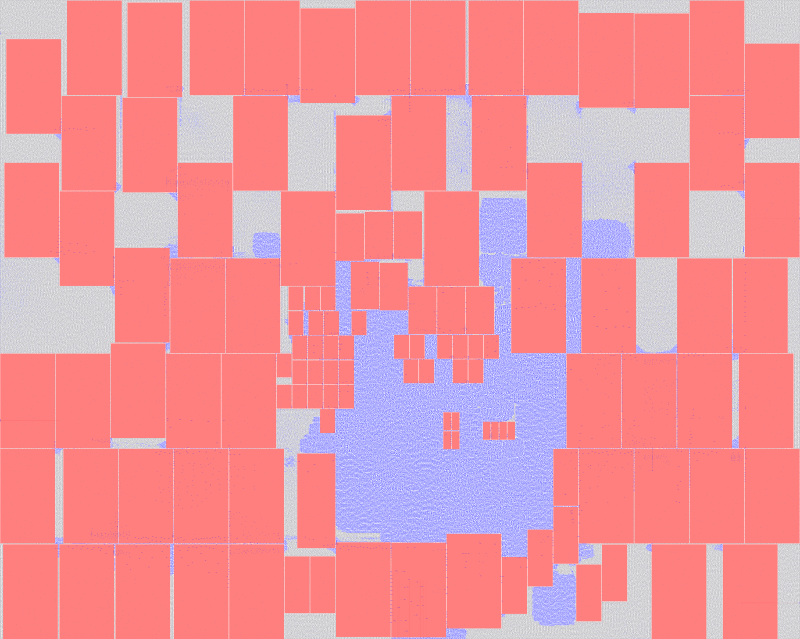}
    }&
    \subfloat[MacroAgent]{%
      \centering\includegraphics[width=.25\linewidth]{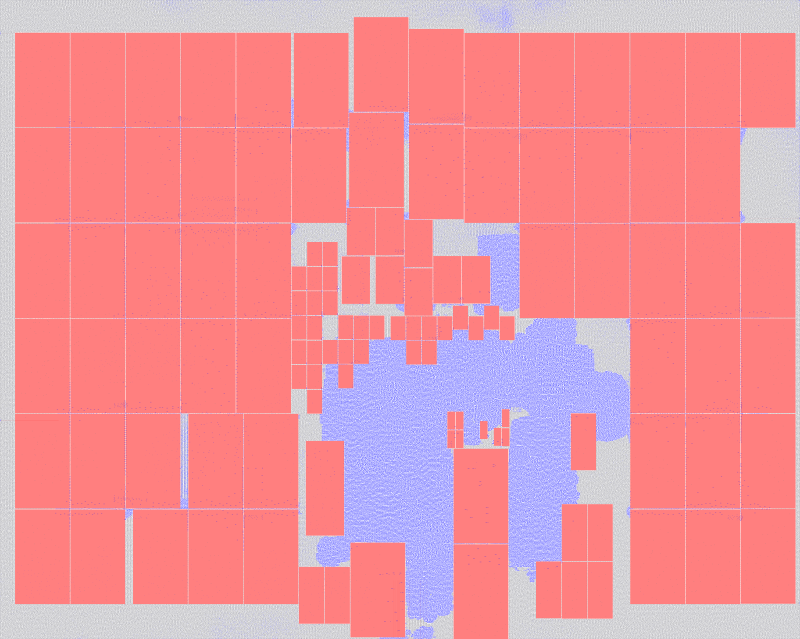}
    }
    \end{tabular}
  \endgroup
  \caption{\revise{Comparison of different macro legalization methods on \texttt{rocket}: (a) initial global placement, (b) DREAMPlace legalization with remaining overlaps, (c) Sequence Pair legalization with irregular channels, and (d) MacroAgent legalization with overlap-free and regular layout.}}
  \label{fig:viz_rocket}
\end{figure}

To intuitively understand the impact of our MacroAgent legalization, we choose \texttt{rocket} to visualize the results of each legalization method.
The visualization results of \texttt{rocket} are shown in \Cref{fig:viz_rocket}. Initial shows the layout
after mixed-size placement, which is the input of macro legalization. 
We can see that the layout after DREAMPlace macro legalization still has overlaps, which shows that one or two \revise{heuristics} cannot cover all testcases.
Although sequence pair can resolve the overlaps, the regularity is not satisfactory, resulting in irregular channels, which is harmful to wirelength.
On the last figure, MacroAgent shows that the layout resolves the overlaps, and the regularity is significantly improved.

\begin{figure}[tb]
    \centering
    \begin{tikzpicture}
        \begin{axis}[
            width=\linewidth,
            height=4.2cm,
            xlabel={Number of Templates },
            ylabel={Wirelength Ratio},
            xtick={1,2,3,4,5},
            xticklabels={1,3,5,7,9},
            ymin=0.95, ymax=1.01,
            ytick={0.96,0.97,0.98,0.99,1.00},
            yticklabel style={/pgf/number format/.cd, fixed, precision=3},
            grid=major,
            every major tick/.append style={
                major tick length=2pt,black},
            every minor tick/.append style={
                minor tick length=1.5pt,gray},
            nodes near coords,
            nodes near coords style={
                /pgf/number format/assume math mode,
                /pgf/number format/fixed,
                /pgf/number format/precision=3, font=\footnotesize, anchor=south},
            x tick label style={
              /pgf/number format/assume math mode,
              /pgf/number format/1000 sep={}},
            y tick label style={
              /pgf/number format/assume math mode,
              /pgf/number format/1000 sep={}},
            every axis plot/.append style={thick, mark=*},
            tick label style={font=\footnotesize},
            label style={font=\footnotesize},
        ]
        \addplot+[color={rgb:red,225;green,235;blue,246}] coordinates {(1,1.000) (2,0.974) (3,0.970) (4,0.967) (5,0.962)};
        \end{axis}
    \end{tikzpicture}
    \caption{The impact of the used contour algorithm number on the average routed wirelength ratio in the TILOS benchmark (starts with only a rectangle contour).}
    \label{fig:hpwl-ratio-avg-line}
\end{figure}
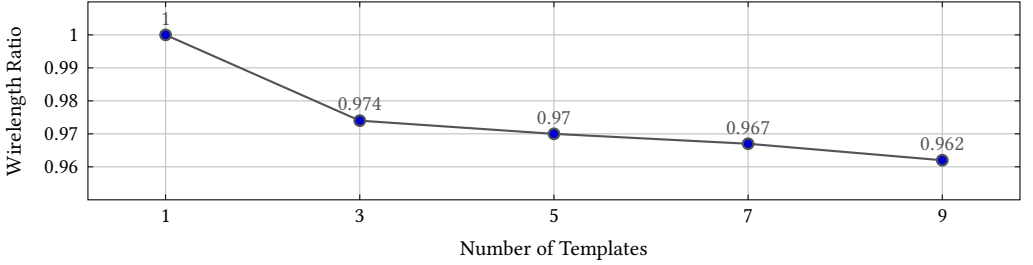

To further demonstrate the effectiveness of LLM-generated multiple contour algorithms, we present the wirelength ratio (same \revise{configuration} with the main experiment) on the TILOS Benchmark under two scenarios: simply using \revise{the} rectangle contour algorithm and continuously adding LLM-generated algorithms. The x-axis counts the number of distinct LLM-designed contour algorithms included in the candidate set (starting from the rectangle baseline = 1). Each contour induces one template; we then pick the best legalization per cluster via assignment. We can see that wirelength is improved when multiple LLM-generated algorithms are used, as shown in \Cref{fig:hpwl-ratio-avg-line}, which shows that LLM-generated algorithms are better and multiple heuristics can help improve the overall performance. 

\subsection{Discussions}
Previous work~\cite{liu2024systematic} indicate that LLMs exhibit limitations in both domain-specific EDA knowledge and large-scale project handling capabilities, 
resulting in suboptimal direct code generation. To address these challenges, we developed a framework that abstracts macro legalization into a geometric  
problem. Notably, when generating algorithms for this abstracted problem, LLMs successfully leveraged universal geometric concepts such as alpha shapes, 
k-nearest neighbors, and minimum spanning trees, rather than relying on EDA-specific domain knowledge. 
\revise{Importantly, our framework employs a general-purpose, off-the-shelf LLM without any domain-specific fine-tuning. Domain-specific decisions---such as the clustering strategy and the choice of optimization objectives (regularity and displacement minimization)---are encoded by human engineers in the framework design, while the LLM operates solely on the abstracted geometric subproblem using its pretrained reasoning capabilities.}
This domain-agnostic approach yielded promising results.

A traditional automation loop optimizes parameters within a fixed algorithm. In contrast, the LLM Agent demonstrated the ability to traverse the algorithmic search space, proposing distinct geometric heuristics (e.g., switching from convex hulls to MST-based contours) that a parameter-sweep baseline could not discover.

Our algorithm enforces regularity solely within intra-cluster legalization, 
while the fallback heuristic algorithms for inter-cluster legalization compromise layout regularity. Meanwhile, the current LLM Agent-based algorithm design approach remains limited to generating contour algorithms. Future research should prioritize advanced regularity-aware heuristic methodologies for clustering and inter-cluster optimization.

\section{Conclusion}
\label{sec:conclusion}


We introduce MacroAgent, a robust regularity-aware macro legalization framework utilizing LLM-agent-designed contour algorithms.
The framework proceeds through four stages: clustering, regularity-aware contour generation, template-based matching, and inter-cluster refinement.
We abstract the EDA problem of macro legalization into a domain-agnostic geometric problem, 
enabling LLMs to design diverse efficient heuristic contour algorithms for macro legalization. 
On TILOS benchmarks, MacroAgent reduces wirelength by 5\% compared to DREAMPlace 2.0~\cite{lin2020dreamplace} and 4\% compared to the sequence pair~\cite{chen2023stronger}, while matching their congestion qualities. 
On Chipyard designs, it successfully legalizes every case and achieves wirelength improvements of 3\% to 5\%.
\revise{End-to-end evaluation through Cadence Innovus further confirms that the regularity improvements yield tangible PPA gains, including 2.9\% lower routed wirelength and 68.3\% TNS improvement over the DREAMPlace macro legalization baseline, and 1.8\% lower routed wirelength when integrated into the Innovus macro placement flow.}
Our approach shortens development cycles and highlights the potential of LLMs to assist in designing EDA algorithms. 
We believe this paradigm can be extended to other NP-hard EDA algorithm design problems, bringing new vitality to the  community.

\balance
{
    \bibliographystyle{IEEEtran}
    \bibliography{ref/Top,ref/All}
}

\end{document}